\pdfoutput=1
\PassOptionsToPackage{table}{xcolor}
\documentclass[sigconf,screen]{acmart}
 
\usepackage{adjustbox}

\usepackage{multirow}
\usepackage{dblfloatfix}
\usepackage{enumitem}
\usepackage{bigstrut}
\usepackage[tikz]{bclogo}
\usepackage[skins]{tcolorbox}
\usepackage{wrapfig}
\usepackage[flushleft]{threeparttable}
\usepackage{float}
 
\definecolor{deeppink}{HTML}{D28986} 
\definecolor{pink}{HTML}{FFABA7} 
\definecolor{lightpink}{HTML}{FFCCC9} 
\usepackage[usestackEOL]{stackengine}
\usepackage{xcolor}

\usepackage[final]{listings}
\usepackage[linesnumbered,ruled,vlined]{algorithm2e}

\SetCommentSty{mycommfont}

\SetKwInput{KwInput}{Input}                
\SetKwInput{KwOutput}{Output}              

\SetKwFunction{FMain}{Main}
\SetKwFunction{FSum}{Sum}
\SetKwFunction{FSub}{get\_ngbr}

\AtBeginDocument{%
  \providecommand\BibTeX{{%
    \normalfont B\kern-0.5em{\scshape i\kern-0.25em b}\kern-0.8em\TeX}}}

\newcommand{\IT}[1]{{\bf%
DODGE(\ifx*#1$\mathcal{E}$\else#1\fi)}}

\newcommand{\bi}{\begin{itemize}[leftmargin=0.4cm]}
	\newcommand{\ei}{\end{itemize}}
\newcommand{\be}{\begin{enumerate}[leftmargin=0.4cm]}
	\newcommand{\ee}{\end{enumerate}}

\begin{document}
\title{Bias in Machine Learning Software: Why? How? What to Do?}
 
\author{Joymallya Chakraborty}
\email{jchakra@ncsu.edu}
\affiliation{%
\institution{North Carolina State University}
\city{Raleigh}
\country{USA}} 

\author{Suvodeep Majumder}
\email{smajumd3@ncsu.edu}
\affiliation{%
\institution{North Carolina State University}
\city{Raleigh}
\country{USA}}

\author{Tim Menzies}
\email{timm@ieee.org}
\affiliation{%
\institution{North Carolina State University}
\city{Raleigh}
\country{USA}}

\begin{abstract}

Increasingly, software is making autonomous decisions in case of criminal sentencing, approving credit cards, hiring employees, and so on. Some of these decisions show bias and adversely affect certain social groups (e.g. those defined by sex, race, age, marital status). Many prior works on bias mitigation take the following form: change the data or learners in multiple ways, then see if any of that improves fairness. Perhaps a better approach is to postulate root causes of bias and then applying some resolution strategy.

This paper checks if the root causes of bias are the prior decisions about (a) what data was selected and (b) the labels assigned to those examples. Our Fair-SMOTE algorithm removes biased labels; and rebalances internal distributions so that, based on sensitive attribute,  examples are equal in  positive and negative classes. On testing,   this method was just as effective at reducing bias as prior approaches. Further, models generated via Fair-SMOTE achieve higher performance (measured in terms of recall and F1) than other state-of-the-art fairness improvement algorithms.

To the best of our knowledge, measured in terms of number of analyzed learners and datasets,
this study is one of the largest studies on bias mitigation yet presented in the literature.

\end{abstract}



\setcopyright{acmcopyright}
\acmPrice{15.00}
\acmDOI{10.1145/3468264.3468537}
\acmYear{2021}
\copyrightyear{2021}
\acmSubmissionID{fse21main-p54-p}
\acmISBN{978-1-4503-8562-6/21/08}
\acmConference[ESEC/FSE '21]{Proceedings of the 29th ACM Joint European Software Engineering Conference and Symposium on the Foundations of Software Engineering}{August 23--28, 2021}{Athens, Greece}
\acmBooktitle{Proceedings of the 29th ACM Joint European Software Engineering Conference and Symposium on the Foundations of Software Engineering (ESEC/FSE '21), August 23--28, 2021, Athens, Greece}

\maketitle

\section{Introduction}

It is the ethical duty of software researchers and engineers to produce  quality software that makes fair decisions, especially  for   high-stake software that makes important decisions about  human lives. 
Sadly, there are   too many examples of machine learning
software  exhibiting unfair/biased behavior based on some protected attributes like sex, race, age, marital status. 
For example:
\bi
\item Amazon had to scrap an automated recruiting tool as it was found to be biased against women \cite{Amazon_Recruit}.
\item
A widely used face recognition software was found to be biased against dark-skinned women \cite{Skin_Bias}.
\item
Google Translate, the most popular translation engine in the world, shows gender bias. ``She is an engineer, He is a nurse'' is translated into Turkish and then again into English becomes ``He is an engineer, She is a nurse'' \cite{Caliskan183}. 
\ei
Prior works on managing fairness in machine learning software have tried mitigating bias by exploring a very wide  space of control parameters for machine learners.
For example,  Johnson et al.~\cite{johnson2020fairkit} executed
some
(very long) grid search that looped over the set of  control parameters of a learner to find settings that reduced bias. Another approach, published in  
FSE'20 by Chakraborty et al. \cite{Chakraborty_2020} used stochastic sampling to explore the space of control options (using some
incremental feedback operator to adjust where to search next).
While these approaches were certainly useful in the domains
tested by Johnson and Chakraborty et al., 
these methods are ``dumb'' in a way because they do not take advantage of domain knowledge.  This is a problem since,
as shown by this paper, such domain knowledge can lead to a more direct (and  faster and more effective) bias mitigation strategy.

The   insight we offer here is that
the root causes of bias might be the {\em prior decisions  that generated the training data}. Those prior decisions affect
(a)~what data was collected and (b)~the labels assigned to those examples. Hence, to fix bias, it might suffice to apply mutators to the training data in order to
\begin{quote}
{\em 
(A)~remove biased labels; and (B)~rebalance  internal distributions such that they are equal based on class and sensitive attributes.}
\end{quote}
This ``Fair-SMOTE'' tactic
uses {\em situation testing}~\cite{USA_SituationTesting,EU_SituationTesting} to find biased labels. Also, it  balances frequencies of  sensitive attributes and class labels
(whereas the older SMOTE algorithm~\cite{Chawla_2002} just balances the class labels).  We recommend Fair-SMOTE since:
\bi
\item
Fair-SMOTE is   {\em  faster} 
(220\%) than prior methods such as \sloppy{Chakraborty} et al.'s~\cite{Chakraborty_2020} method.
\item Models generated via  Fair-SMOTE  are more {\em effective} (measured in terms of recall and F1) than
Chakraborty et al. \cite{Chakraborty_2020}
and another state-of-the-art bias mitigation algorithm~\cite{NIPS2017_6988}.
\ei
Overall, this paper makes the following contributions:

\begin{figure*}[!t]
\label{Confusion}
\centering
\includegraphics[width=.8\linewidth]{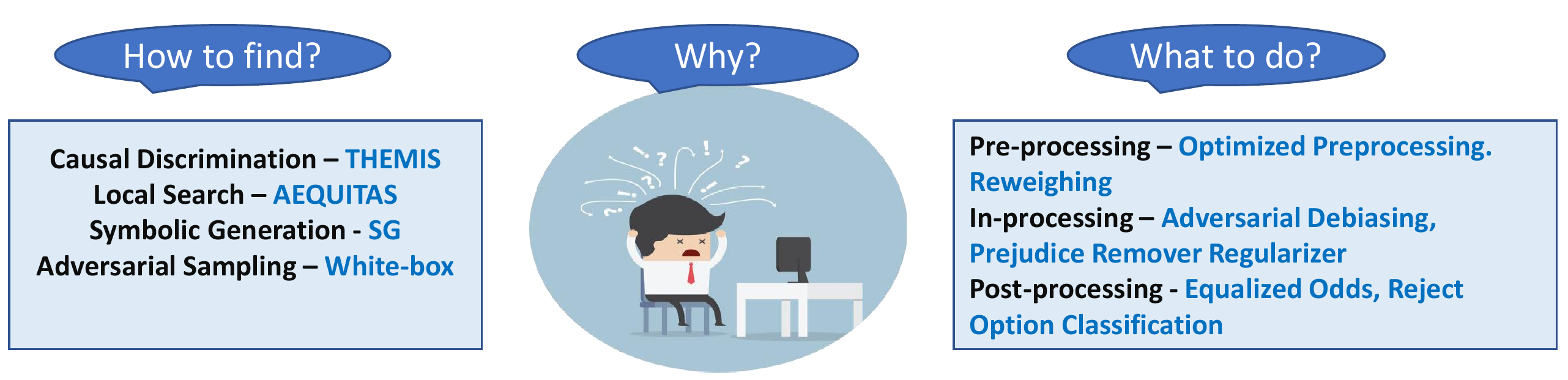}
\caption{Many  tools try to find  or explain or  mitigate bias. Fair-SMOTE addresses all three problems, at the same time.}
\end{figure*}

\begin{itemize}
    \item We demonstrate two main reasons for the training data being biased - a) data imbalance, b) improper data label.
    \item We combine finding bias, explaining bias, and removing bias.
    \item We show that traditional class balancing techniques damage the fairness of the model.
    \item Prior works compromised performance while achieving fairness. We achieve fairness with better recall and F1 score.
    \item To the best of our knowledge, this study explores more learners and datasets than prior works that were based on just a handful of datasets and/or learners (e.g.~\cite{NIPS2017_6988,zhang2018mitigating,Kamiran:2018:ERO:3165328.3165686}).
\end{itemize}

Before beginning, we digress to clarify two points. Firstly, in this paper, {\em training data} is mutated by Fair-SMOTE but the {\em test data} remains in its original form. 

Secondly,  one danger with mutating data is that important associations between variables can be lost. Hence, in this work, we take care to mutate by extrapolating between the values seen in two neighboring examples. In that mutation process, Fair-SMOTE {\em extrapolates all the variables by the same amount}. That is, if there exists some average case association between the pre-mutated examples, then that association is preserved in the mutant. To facilitate open science, the source code and datasets used in this study are all available online\footnote{https://github.com/joymallyac/Fair-SMOTE}.

\section{Related Work}
\label{previous_work}

Fairness in ML model is a well-explored research topic in the ML Community. Recently, the software engineering community has also become interested in the problem of fairness.  Software researchers from UMass Amherst have developed a python toolkit called Fairkit to evaluate ML models based on various fairness metrics \cite{johnson2020fairkit}. ICSE 2018 and ASE 2019 conducted separate workshops for software fairness \cite{FAIRWARE,EXPLAIN}. Big software industries have started taking this fairness problem seriously as well. IBM has created a public GitHub repository AI Fairness 360 \cite{AIF360} where most popular fairness metrics, and mitigation algorithms can be found. Microsoft has created a dedicated research group for fairness, accountability, transparency, and ethics in AI \cite{FATE}. Facebook has developed a tool to detect bias in their internal software \cite{Fairness_Flow}.

\noindent
This section reviews the fairness   literature in order to find  two baseline systems (which we will use to  evaluate Fair-SMOTE). 

Our first baseline is  Chakraborty et al.~\cite{Chakraborty_2020} from FSE'20 that viewed  bias mitigation as a multiple-goal  hyperparameter optimization problem. Their stochastic search found learner control  parameters that reduced prediction bias. 
We prefer  Chakraborty et al.'s method over the grid search of Johnson et al.~\cite{johnson2020fairkit}   since grid search (a)~can be very slow and (b)~it is strongly deprecated in the machine learning literature~\cite{bergstra2012random}.
That said, one drawback with the method of Chakraborty et al. is that it did not guide its search via domain knowledge (such as those listed in the introduction).

\begin{table*}[]
\caption{Description of the datasets used in the experiment.}
\label{Datasets}
\small
\begin{tabular}{|c|c|c|c|l|}
\hline
\rowcolor[HTML]{C0C0C0} 
Dataset             & \#Rows   & \#Features & \begin{tabular}[c]{@{}c@{}}Protected \\ Attribute\end{tabular} & \multicolumn{1}{c|}{\cellcolor[HTML]{C0C0C0}Description}                                       \\ \hline
Adult Census Income \cite{ADULT} & 48,842   & 14         & Sex, Race                                                      & Individual information from 1994 U.S. census. Goal is predicting income \textgreater \$50,000. \\ \hline
Compas \cite{COMPAS}             & 7,214    & 28         & Sex,Race                                                       & Contains criminal history of defendants. Goal predicting re-offending in future                \\ \hline
German Credit \cite{GERMAN}      & 1,000    & 20         & Sex                                                            & Personal information about individuals \& predicts good or bad credit.                         \\ \hline
Default Credit \cite{DEFAULT}     & 30,000   & 23         & Sex                                                            & Customer information for people from Taiwan. Goal is predicting default payment.               \\ \hline
Heart Health  \cite{HEART}      & 297      & 14         & Age                                                            & Patient information from Cleveland DB. Goal is predicting heart disease.                       \\ \hline
Bank Marketing \cite{BANK}     & 45,211   & 16         & Age                                                            & Contains marketing data of a Portuguese bank. Goal predicting term deposit.                    \\ \hline
Home Credit \cite{HOME_CREDIT}        & 37,511 & 240        & Sex                                                            & Loan applications for individuals. Goal is predicting application accept/reject.               \\ \hline
Student Performance \cite{STUDENT} & 1,044    & 33         & Sex                                                            & Student achievement of two Portuguese schools. Target is final year grade.                     \\ \hline
MEPS-15,16  \cite{MEPS}        & 35,428   & 1,831      & Race                                                           & Surveys of families, individuals, medical providers, employers. Target is ``Utilization''.     \\ \hline
\end{tabular}
\end{table*}

\begin{table*}[]
\small
\caption{Definition of the performance and fairness metrics used in this study.}
\label{metrics_table}
\begin{tabular}{|l|c|l|c|}
\hline
\rowcolor[HTML]{C0C0C0} 
\multicolumn{1}{|c|}{\cellcolor[HTML]{C0C0C0}{\color[HTML]{333333} \textbf{Performance Metric}}} & {\color[HTML]{333333} \textbf{\begin{tabular}[c]{@{}c@{}}Ideal \\ Value\end{tabular}}} & \multicolumn{1}{c|}{\cellcolor[HTML]{C0C0C0}{\color[HTML]{333333} \textbf{Fairness Metric}}}                                                                                                                                                                                                                                                                                                                       & {\color[HTML]{333333} \textbf{\begin{tabular}[c]{@{}c@{}}Ideal \\ Value\end{tabular}}} \\ \hline
Recall = TP/P = TP/(TP+FN)                                                                       & 1                                                                                      & \begin{tabular}[c]{@{}l@{}}\textbf{Average Odds Difference (AOD)}: Average of difference in False Positive Rates(FPR) and \\ True Positive Rates(TPR) for unprivileged and privileged groups \cite{IBM}. \\ TPR = TP/(TP + FN),  FPR = FP/(FP + TN), $AOD = {[}(FPR_{U} - FPR_{P}) + (TPR_{U} - TPR_{P}){]} * 0.5$\end{tabular}                                                                                                             & 0                                                                                      \\ \hline
False alarm = FP/N = FP/(FP+TN)                                                                  & 0                                                                                      & \begin{tabular}[c]{@{}l@{}}\textbf{Equal Opportunity Difference (EOD)}:  Difference of True Positive Rates(TPR) for \\ unprivileged and privileged groups \cite{IBM}. $EOD = TPR_{U} - TPR_{P}$\end{tabular}                                                                                                                                                                                                                               & 0                                                                                      \\ \hline
Accuracy = $ \frac {(TP + TN)}{(TP + FP + TN + FN)}$                                                         & 1                                                                                      & \begin{tabular}[c]{@{}l@{}}\textbf{Statistical Parity Difference (SPD)}: Difference between probability  of unprivileged \\ group (protected attribute PA = 0) gets favorable prediction ($\hat{Y} = 1$) \& probability \\ of privileged group (protected attribute PA = 1) gets favorable prediction ($\hat{Y} = 1$) \cite{10.1007/s10618-010-0190-x}.\\ $SPD = P[\hat{Y}=1|PA=0] - P[\hat{Y}=1|PA=1]$\end{tabular} & 0                                                                                      \\ \hline
Precision = TP/(TP+FP)                                                                           & 1                                                                                      & \begin{tabular}[c]{@{}l@{}}\textbf{Disparate Impact (DI)}: Similar to SPD but instead of the difference of probabilities, \\ the ratio is measured \cite{feldman2015certifying}. $DI = P[\hat{Y}=1|PA=0]/P[\hat{Y}=1|PA=1]$\end{tabular}                                                                                                                                                                         & 1                                                                                      \\ \hline
F1 Score = $ \frac {2 \;*\; (\mathit{Precision} \;*\; \mathit{Recall}) }{(\mathit{Precision}\; + \; \mathit{Recall})}$                            & 1                                                                                      &                                                                                                                                                                                                                                                                                                                                                                                                                    &                                                                                        \\ \hline
\end{tabular}
\end{table*}

For a second baseline system, we reviewed popular bias mitigation works including Reweighing ~\cite{Kamiran2012}, Prejudice Remover Regularizer~ \cite{10.1007/978-3-642-33486-3_3}, Adversarial debiasing~ \cite{zhang2018mitigating}, Equality of Opportunity~\cite{hardt2016equality}, and Reject option classification~ \cite{Kamiran:2018:ERO:3165328.3165686}.
While all these tools are useful for mitigating bias,
as a side effect of {\em improving} fairness these methods also {\em degrade} learner performance (as measured by accuracy \& F1~\cite{NIPS2017_6988,Kamiran2012,10.1007/978-3-642-33486-3_3,zhang2018mitigating,Kamiran:2018:ERO:3165328.3165686,hardt2016equality}).
Hence,  a truism in the research community is that, as said by Berk et al.~\cite{berk2017fairness}:
\begin{quote}
{\em It is impossible to achieve fairness and high performance simultaneously (except in trivial cases).}
\end{quote}
In 2020, Maity et al. tried to reverse the Berk et al. conclusion. To do so they had to make the unlikely assumption that the test data was unbiased~\cite{maity2020tradeoff}. 
Our experience is that this assumption is difficult to achieve. Also,
based on our experience with Fair-SMOTE (described below),
we argue that this assumption may not even be necessary. 
We show below that it is possible to mitigate bias while maintaining/improving performance at the same time. The trick is how the data is mutated.  Fair-SMOTE {\em extrapolates all the variables by the same amount}. In this way,   associations between variables (in the pre-mutated examples), can be preserved in the mutant.

 As to other work, a recent paper from Google \cite{jiang2019identifying} talked about data labels being biased. This motivated us to look into data labels. Yan et al. commented that under representation of any sensitive group in the training data may give too little information to the classifier resulting bias \cite{10.1145/3340531.3411980}. This work inspired us to look into data distribution and class distribution to find root causes for bias.
 
From the remaining  literature, we  would characterize much of it as  looking {\em vertically} down a few columns of data (while in our approach, 
 our mutators work {\em horizontally} across rows of data). Further,  much prior works only  {\em test} for bias (whereas in our approach, we propose how to {\em fix} that bias). 
 
 There are many examples  of ``vertical'' studies that just test for bias. Zhang et al.~\cite{zhang2016causal} built a directed acyclic graph (DAG) containing cause-effect relations of all the attributes to see which features directly or indirectly affect the outcome of the model. Loftus et al.~\cite{loftus2018causal} show methods to detect if  sensitive attributes such as race, gender are implicated in model outcome.
   THEMIS~\cite{Galhotra_2017}  randomly perturbs each attribute to see whether model discriminates amongst individuals. AEQUITAS~\cite{Udeshi_2018},
  uses random, semi-directed, and fully-directed generation techniques that make it more efficient than THEMIS. 
  Recently, IBM researchers  have developed a new testing technique by combining LIME (Local Interpretable Model Explanation) and symbolic execution~\cite{Aggarwal:2019:BBF:3338906.3338937}. Zhang et al. developed a fairness testing method that is applicable for DNN models~\cite{10.1145/3377811.3380331} (they combined gradient computation and clustering techniques).

Finally, we can find one paper that seems like a competitive technology to the methods of this paper.
 Optimized Pre-processing (OP) by Calmon et al.~\cite{NIPS2017_6988} tries to find, then fix,  bias. OP is a data-driven optimization framework for probabilistically transforming data in order to reduce algorithmic discrimination. OP treats bias mitigation as a convex optimization problem with goals to preserve performance and achieve fairness. We chose this work as a baseline because, like Fair-SMOTE, it is also a pre-processing strategy.  

In summary, prior work suffered from
\bi
\item Some methods only  find bias, without trying to fix it. 
\item
Some methods for fixing bias have an undesired side effect: leaner performance was degraded. 
\ei
For the rest of this paper, we explore a  solution that  finds root causes of bias, and directly implement mitigation of those causes (resulting in less bias and better performance than seen in prior work).

\section{What are the Root Causes of Bias?}
\label{root_cause}
To understand what might cause bias, at first, we need some terminology. Table \ref{Datasets} contains 10 datasets  used in this study. All of them are binary classification problems where the target class has only two values.
\begin{itemize}
\item
A class label is called a \textit{favorable label} if it gives an advantage to the receiver such as receiving a loan, being hired for a job.
\item
A \textit{protected attribute} is an attribute that divides the whole population into two groups (privileged \& unprivileged) that have differences in terms of   receiving benefits. 
\end{itemize}
  Every dataset in Table~\ref{Datasets} has one or two such attributes. For example, in case of credit card application datasets, based on protected attribute ``sex'', ``male'' is privileged and ``female'' is unprivileged; in case of health datasets, based on protected attribute ``age'', ``young'' is privileged and ``old'' is unprivileged.
\bi
\item
\textit{Group fairness} is the goal that based on the protected attribute, privileged and unprivileged groups will be treated similarly.
\item
\textit{Individual fairness} means similar outcomes go to similar individuals. 
\ei

\begin{figure*}
\includegraphics[width=\textwidth]{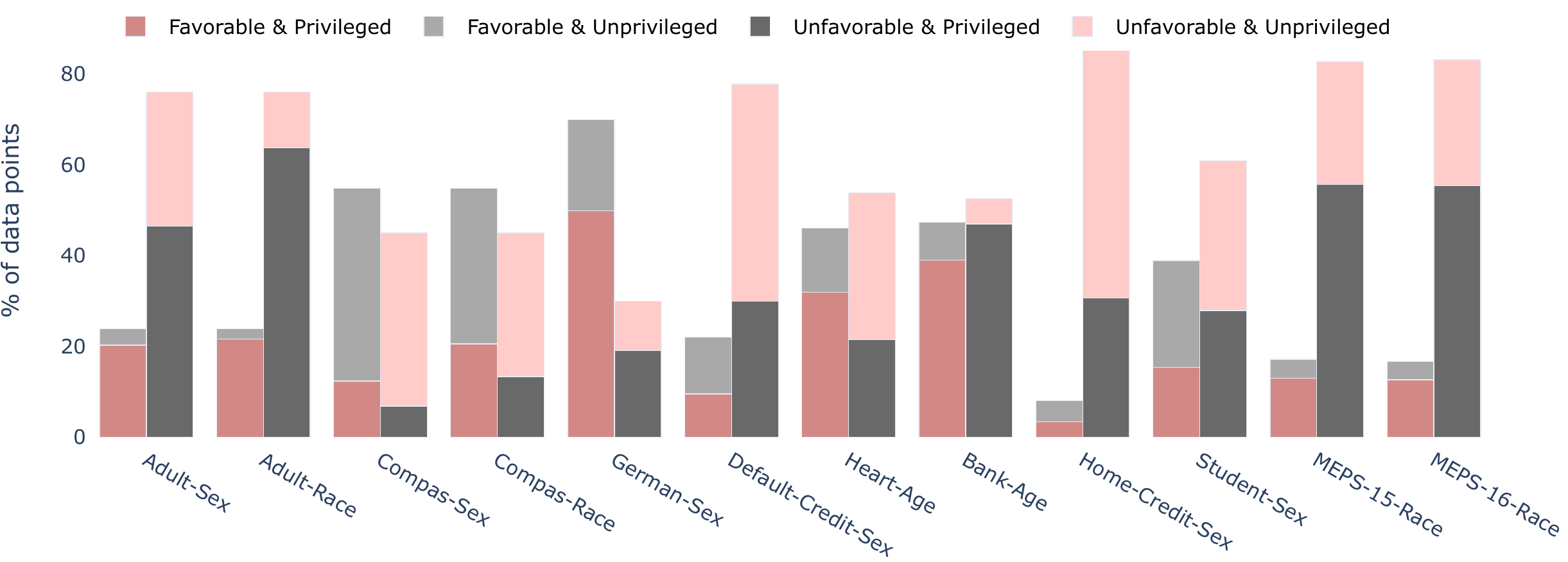}
\caption{Most of the datasets showing not only class imbalance but also imbalance based on the protected attribute.}
\label{Class_Protected_Imbalance}
\end{figure*}

\begin{figure}
\includegraphics[width=0.45\textwidth]{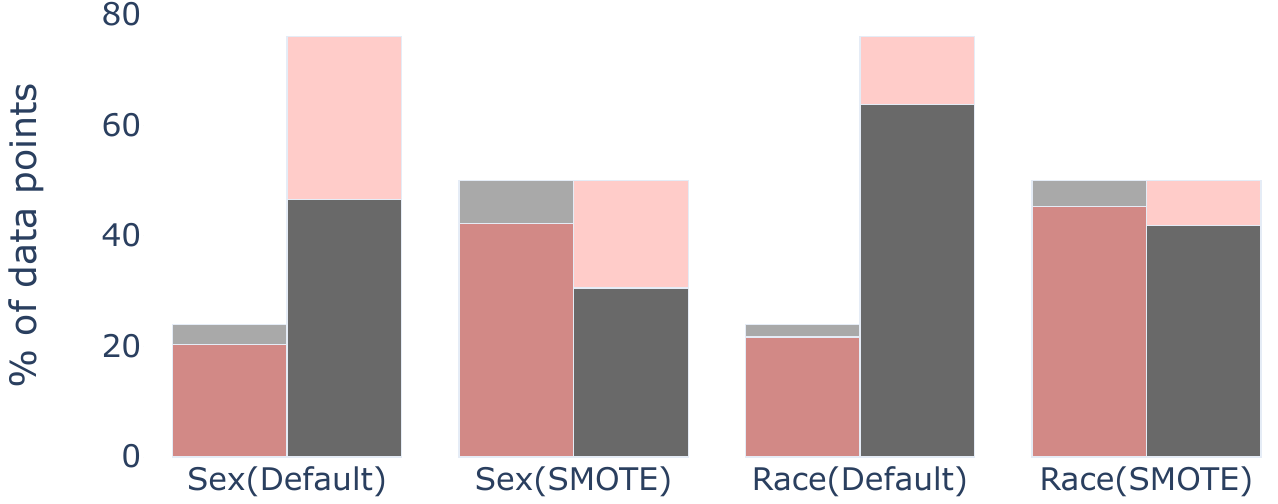}
\caption{Effects of SMOTE class balancing technique on Adult dataset for two protected attributes ``sex'' and ``race''.}
\label{Adult_Imbalance}
\end{figure}

Table \ref{metrics_table} contains five performance metrics and four fairness metrics we used. 
These metrics are selected since they were widely used in the literature~\cite{Chakraborty_2020,Biswas_2020,chakraborty2019software,hardt2016equality,9286091}.
Prediction performance is measured in terms of
{\em recall, false alarm, accuracy, precision, F1};
fairness is measured using
{\em AOD, EOD, SPD, DI}.
All of these can be calculated from the confusion matrix of binary classification containing four cells - true positives (TP), false positives (FP), true negatives (TN) and false negatives (FN). 
\begin{itemize}
\item For recall, accuracy, precision, F1 {\em larger values} are {\em better};
\item For false alarm, AOD, EOD, SPD {\em smaller values} are {\em better}.
\item
DI is a ratio and there is no bias when value of DI is 1. For comprehensibility, while showing results we compute \mbox{abs(1 - DI)} so that all four fairness metrics are lower the better (0 means no bias).
\end{itemize}
Prior research \cite{Kamiran2012,Angell:2018:TAT:3236024.3264590,Kamiran:2018:ERO:3165328.3165686,10.1007/978-3-642-33486-3_3} has shown that classification models built from the datasets of Table~\ref{Datasets} show bias. Why? What are the natures of these datasets that result in bias? 

In this paper, we postulate that data has a history and that history introduces bias.
For example, consider a team of humans labeling data as ``risky loan applicant'' or otherwise. If that team has any biases (conscious or otherwise) against certain social groups, then the bias of those humans introduces improper/unfair data labels.

Another way history can affect the data is {\em selection bias}. Suppose we are collecting package data relating to the kinds of items shipped via Amazon to a particular suburb.
Suppose there is some institutional bias that tends to results in low annual incomes for persons in certain suburbs\footnote{ E.g. If politicians    spend less money on  schools in a poorer district; then
that district has (a)~fewer exceptional schools; 
(b)~fewer graduates with job skills in  for high-paying jobs in high demand;
(c)~consistently lower incomes from generation
to generation.}. The delivery data collected from those suburbs would contain all the political and economic biases that tend to perpetuate lower socio-economic neighborhoods.

The next two sections explore the intuitions of the last two paragraphs in more detail. Specifically, we will look into data imbalance and improper data labeling. This in turn will lead to the Fair-SMOTE algorithm that deletes biased labels and balances all distributions between positive and negative examples.

\begin{table*}[!t]
\small
\caption{Effects of various class balancing techniques on Adult dataset (note: for the metrics with `+' more is better and for the metrics with `-' less is better). For each metric, cell with best score is highlighted.}
\label{Sampling_Adult}
\begin{tabular}{|c|c|c|c|c|c|c|c|c|c|}
\hline
\rowcolor[HTML]{C0C0C0} 
\textbf{}                                                  & Recall(+)                                           & False alarm(-)                                      & Precision(+)                                        & Accuracy(+)                                         & F1 Score(+)                                         & AOD(-)                                              & EOD(-)                                              & SPD(-)                                              & DI(-)                                               \\ \hline
\cellcolor[HTML]{C0C0C0}Default                            & {\color[HTML]{000000} 0.42}                         & \cellcolor[HTML]{FFABA7}{\color[HTML]{000000} 0.07} & \cellcolor[HTML]{FFABA7}{\color[HTML]{000000} 0.69} & \cellcolor[HTML]{FFABA7}{\color[HTML]{000000} 0.83} & {\color[HTML]{000000} 0.54}                         & {\color[HTML]{000000} 0.12}                         & \cellcolor[HTML]{FFABA7}{\color[HTML]{000000} 0.24} & \cellcolor[HTML]{FFABA7}{\color[HTML]{000000} 0.21} & \cellcolor[HTML]{FFABA7}{\color[HTML]{000000} 0.56} \\ \hline
\cellcolor[HTML]{C0C0C0}RUS                                & \cellcolor[HTML]{FFABA7}{\color[HTML]{000000} 0.74} & {\color[HTML]{000000} 0.23}                         & {\color[HTML]{000000} 0.48}                         & {\color[HTML]{000000} 0.76}                         & {\color[HTML]{000000} 0.59}                         & {\color[HTML]{000000} 0.09}                         & {\color[HTML]{000000} 0.36}                         & {\color[HTML]{000000} 0.37}                         & {\color[HTML]{000000} 0.61}                         \\ \hline
\cellcolor[HTML]{C0C0C0}ROS                                & \cellcolor[HTML]{FFABA7}{\color[HTML]{000000} 0.74} & {\color[HTML]{000000} 0.24}                         & {\color[HTML]{000000} 0.47}                         & {\color[HTML]{000000} 0.75}                         & {\color[HTML]{000000} 0.59}                         & \cellcolor[HTML]{FFABA7}{\color[HTML]{000000} 0.08} & {\color[HTML]{000000} 0.34}                         & {\color[HTML]{000000} 0.35}                         & {\color[HTML]{000000} 0.64}                         \\ \hline
\cellcolor[HTML]{C0C0C0}SMOTE                              & {\color[HTML]{000000} 0.70}                         & {\color[HTML]{000000} 0.25}                         & {\color[HTML]{000000} 0.49}                         & {\color[HTML]{000000} 0.70}                         & \cellcolor[HTML]{FFABA7}{\color[HTML]{000000} 0.64} & {\color[HTML]{000000} 0.17}                         & {\color[HTML]{000000} 0.37}                         & {\color[HTML]{000000} 0.33} & {\color[HTML]{000000} 0.58}                         \\ \hline
\multicolumn{1}{|l|}{\cellcolor[HTML]{C0C0C0}KMeans-SMOTE} & {\color[HTML]{000000} 0.73}                         & {\color[HTML]{000000} 0.24}                         & {\color[HTML]{000000} 0.48}                         & {\color[HTML]{000000} 0.74}                         & {\color[HTML]{000000} 0.58}                         & \cellcolor[HTML]{FFABA7}{\color[HTML]{000000} 0.08} & {\color[HTML]{000000} 0.35}                         & {\color[HTML]{000000} 0.36}                         & {\color[HTML]{000000} 0.62}                         \\ \hline
\end{tabular}
\end{table*}

\begin{table*}[!t]
\caption{Percentage of data points failing situation testing for 10 datasets.}
\label{Situation_Testing}
\small
\begin{tabular}{|c|c|c|c|c|c|c|c|c|c|c|c|c|}
\hline
\rowcolor[HTML]{C0C0C0} 
                                                                                    & \begin{tabular}[c]{@{}c@{}}Adult\\ (Sex)\end{tabular} & \begin{tabular}[c]{@{}c@{}}Adult\\ (Race)\end{tabular} & \begin{tabular}[c]{@{}c@{}}Compas\\ (Sex)\end{tabular} & \begin{tabular}[c]{@{}c@{}}Compas\\ (Race)\end{tabular} & \begin{tabular}[c]{@{}c@{}}German\\ (Sex)\end{tabular} & \begin{tabular}[c]{@{}c@{}}Default\\ Credit\\ (Sex)\end{tabular} & \begin{tabular}[c]{@{}c@{}}Heart-\\ Health\\ (Age)\end{tabular} & \begin{tabular}[c]{@{}c@{}}Bank\\ Marketing\\ (Age)\end{tabular} & \begin{tabular}[c]{@{}c@{}}Home\\ Credit\\ (Sex)\end{tabular} & \begin{tabular}[c]{@{}c@{}}Student\\ (Sex)\end{tabular} & \begin{tabular}[c]{@{}c@{}}MEPS-15\\ (Race)\end{tabular} & \begin{tabular}[c]{@{}c@{}}MEPS-16\\ (Race)\end{tabular} \\ \hline
\cellcolor[HTML]{C0C0C0}\begin{tabular}[c]{@{}c@{}}\% of Rows\\ failed\end{tabular} & 11\%                                                  & 3\%                                                    & 18\%                                                   & 8\%                                                     & 6\%                                                    & 6\%                                                              & 8\%                                                             & 19\%                                                             & 17\%                                                          & 4\%                                                     & 4\%                                                      & 4\%                                                      \\ \hline
\end{tabular}
\end{table*}

\subsection{Data Imbalance}
\label{data_imbalance_subsetion}

When a classification model is trained on imbalanced data, the trained model shows bias against a certain group of people. We mention that such imbalances are quite common. Figure \ref{Class_Protected_Imbalance} displays distributions within our datasets. Note that, in most cases, we see a \textit{class imbalance}; i.e the number of observations per class is not equally distributed. Further, we see that {\em the imbalance is not just based on class, but also on protected attribute}. 

For example, consider the  Adult dataset. Here we are predicting the annual income of a person. There are two classes. ``High income'' ($\ge$ \$50,000) which is \textit{favorable label} and ``low income'' ($<$ \$50,000) which is \textit{unfavorable label}. The first grouped bar in Figure \ref{Class_Protected_Imbalance} has two bars grouped together. The first bar is for ``high income'' class (24\%) and the second bar is for ``low income'' (76\%) class. It is clear that the number of instances for ``low income'' is more than three times of instances for ``high income''. This is an example of class imbalance. 
Now, both the bars are subdivided based on \textit{protected attribute} ``sex'' (``male'' and ``female''). For ``high income'' or \textit{favorable label}, 86\% instances are ``male'' and only 14\% are female. For ``low income'' or \textit{unfavorable label}, 60\% instances are ``male'' and 40\% are female. Overall, this dataset contains more examples of ``male'' (privileged) getting  \textit{favorable label} and female (unprivileged) getting \textit{unfavorable label}.

Class imbalance is a widely studied topic in ML domain. There are mainly two ways of balancing imbalanced classes:
\bi
\item
Oversample the minority class;
\item
or undersample the majority class. 
\ei
We want to see how various class balancing techniques affect fairness. In Table \ref{Sampling_Adult} we are showing the results for five most commonly used class balancing techniques on Adult dataset. One of them is undersampling (RUS- Random Under Sampling), other three (ROS- Random Over Sampling, SMOTE- Synthetic Minority Over Sampling Technique \cite{Chawla_2002} and KMeans-SMOTE \cite{Douzas_2018}) are oversampling techniques. Table \ref{Sampling_Adult} shows values of nine metrics - first five of them are performance metrics and last four of them are fairness metrics. We used logistic regression model here.

The important observation here is all four of the class balancing techniques are increasing bias scores mean damaging fairness (lower is better here). To better understand this, see Figure \ref{Adult_Imbalance}. It gives a visualization of using SMOTE~\cite{Chawla_2002} on Adult dataset. SMOTE generates synthetic samples for minority class data points to equalize two classes. Suppose a data point from minority class is denoted as X where $x_{1},x_{2},..,x_{n}$ are the attributes and its nearest neighbor is $X^{'}$ $(x_{1}^{'}$,$x_{2}^{'}$,..,$x_{n}^{'})$. According to SMOTE, a new data point Y $(y_1,y_2,..,y_n)$ is generated by the following formula:

\begin{align*}
    Y = X + rand(0,1) * (X - X^{'})
\end{align*}

SMOTE definitely balances the majority class and minority class but it damages the protected attribute balance even more. Thus Figure \ref{Adult_Imbalance} explains the results of Table \ref{Sampling_Adult}. 

For space reasons, we have shown this with one dataset and one technique (SMOTE) but in other datasets also, we got similar pattern. That said, traditional class balancing methods improve performance of the model but damage fairness (usually). The reason is {\em these techniques randomly generate/discard samples just to equalize two classes and completely ignore the attributes and hence damage the protected attribute balance even more}.

To fix this, we propose Fair-SMOTE. Like SMOTE, Fair-SMOTE will generate synthetic examples. But Fair-SMOTE takes much more care than SMOTE for exactly how it produces new examples.
Specifically, it balances data based on class and sensitive attributes such that privileged and unprivileged groups have an equal amount of positive and negative examples in the training data  
(for more details, see  \S\ref{methodology}).

\subsection{Improper Data Labeling}
\label{subsection:improper_label}
Some prior works \cite{Das2020fairML,jiang2019identifying,Ted_Simons}   argue that improper labeling could be a reason behind bias.  We used the concept of \textit{situation testing} to validate how labeling can affect fairness of the model. \textit{Situation testing} is a research technique used in the legal field~\cite{USA_SituationTesting} where decision makers' candid responses to applicant’s personal characteristics are captured and analyzed. For example:
\bi
\item
A pair of research assistants (a male and a female with almost equivalent qualities) undergo the same procedure, such as applying for a job. 
\item
Now the treatments they get from the decision-maker are analyzed to find discrimination. 
\ei
\textit{Situation testing} as a legal tactic has been widely used both in the United States \cite{USA_SituationTesting} and the European Union \cite{EU_SituationTesting}. Luong et al. \cite{10.1145/2020408.2020488} first used the concept of situation testing in classification problems. They used K-NN approach to find out similar individuals getting different outcomes to find discrimination. Later, Zhang et al. \cite{10.5555/3060832.3061001} used causal bayesian networks to do the same. The core idea of our situation testing is much simpler:
\bi
\item
Flip the value of protected attribute for every data point. 
\item
See whether prediction given by the model changes or not.
\ei
In our implementation,  a logistic regression model is trained first and then all the data points are predicted. After that, the protected attribute value for every data point is flipped (e.g. male to female, white to non-white) and tested again to validate whether model prediction changes or not. If the result changes, that particular data point fails the situation testing.

Table \ref{Situation_Testing} shows the median of ten runs for all the datasets. We see all the datasets more or less contain these kinds of biased data points. That means we have found \textit{biased labels}. Note that, we use logistic regression model for \textit{situation testing} but any other supervised model can be chosen. We picked logistic regression because it is a simple model and can be trained with a very low amount of data (compared to DL models)~\cite{Bujang2018SampleSG}. Choosing a different model may change the outcome a bit. We will explore that in future.

\section{Methodology}
\label{methodology}

Summarizing the above, we postulate that data imbalance and improper labeling are the two main reasons for model bias. Further, we assert  that if we can solve these two problems then final outcome will be a fairer model generating fairer prediction.  This section describes experiments to test that idea.
\vspace{-0.5cm}
\subsection{Fair-SMOTE}
Fair-SMOTE algorithm solves data imbalance. At first, the training data is divided into subgroups based on class and protected attribute. If class and protected attribute both are binary, then there will be 2*2 = 4 subgroups (Favorable \& Privileged, Favorable \& Unprivileged, Unfavorable \& Privileged, Unfavorable \& Unprivileged). Initially, these subgroups are of unequal sizes. 

Fair-SMOTE synthetically generates new data points for all the subgroups except the subgroup having the maximum number of data points. As a result, all subgroups become of equal size (same with the maximum one).

As stated in the introduction, one danger with mutating data is that important associations between variables can be lost. Hence, in this work, we take care to mutate by extrapolating between the values seen in two neighboring examples. In that mutation process, Fair-SMOTE {\em extrapolates all the variables by the same amount}. That is, if there exists some average case association between the pre-mutated examples, then that association is preserved in the mutant. 

For data generation, we use two hyperparameters ``mutation amount'' (f) and ``crossover frequency'' (cr) like \textit{Differential Evolution} \cite{DE}.  They both lie in the range [0, 1]. The first one denotes at which probability the new data point will be different from the parent point (0.8 means 80\% of the times) and the latter one denotes how much different the new data point will be. We have tried with 0.2 (< 50\% probability), 0.5 (= 50\%), \& 0.8 (> 50\%) and got best results with 0.8.

Algorithm \ref{Algo} describes the pseudocode of Fair-SMOTE. It starts with randomly selecting a data point (parent point p) from a subgroup. Then using K-nearest neighbor, two data points (c1, c2) are selected which are closest to p. Next, according to the algorithm described, a new data point is created. Separate logic is used for boolean, symbolic, and numeric columns. 

Fair-SMOTE does not randomly create a new data point. Rather it creates a data point that is very close to the parent point. Thus the generated data points belong to the same data distribution. This process is repeated until all the subgroups become of similar size. 

After applying Fair-SMOTE, the training data contains equal proportion of both classes and the protected attribute.

\begin{algorithm}[!t]
\footnotesize
\DontPrintSemicolon
    \KwInput{Dataset, Protected Attribute(p\_attrs), Class Label(cl)}
    \KwOutput{Balanced Dataset}
    \SetKwProg{Fn}{Def}{:}{}
      \Fn{\FSub{$Dataset$, $knn$}}{
            rand\_sample\_id = random(0, len(Dataset))\;
            parent = Dataset[rand\_sample\_id]\;
            ngbr = knn.kneighbors(parent,2)\;
            c1, c2 = Dataset[ngbr[0]], Dataset[ngbr[1]]\;
            \KwRet parent, c1, c2\;
      }
    count\_groups = get\_count(Dataset, p\_attrs, cl) \;
    
    max\_size = max(count\_groups) \;
    
    cr, f = 0.8, 0.8  (user can pick any value in [0,1]) \;

    
    \For{c in cl}
    {
        \For{attr in p\_attrs}
        {
            sub\_data = Dataset(cl=c $\And$ p\_attrs=attr)\;
            sub\_group\_size = count\_groups[c][attr]\;
            to\_generate = max\_size - sub\_group\_size\;
            knn = NearestNeighbors(sub\_data)\;
            \For{i in range(to\_generate)}
            {
                p,c1,c2 = get\_ngbr(sub\_data, knn)\;
                new\_candidate = []\;
                \For{col in parent\_candidate.columns}
                {  
                    \If{col is boolean}
                    {
                        \If{cr > random(0,1)}
                        {
                            new\_val = random(p[col],c1[col],c2[col])\;
                        }
                        \Else
                        {
                            new\_val = p[col]\;
                        }
                        new\_candidate.add(new\_val)\;
                    }
                    \ElseIf{col is String}
                    {
                        new\_val = random(p[col],c1[col],c2[col])\;
                        new\_candidate.add(new\_val)\;
                    }
                    \ElseIf{col is Numeric}
                    {
                        \If{cr > random(0,1)}
                        {
                            new\_val = p[col] + f*(c1[col]-c2[col])\;
                        }
                        \Else
                        {
                        	new\_val = p[col]\;
                        }
                        new\_candidate.add(new\_val)\;
                    }
                }
            }
            Dataset.add(new\_candidate)
        }
    }
        
\caption{Pseudocode of Fair-SMOTE}
\label{Algo}
\end{algorithm}

\begin{figure}
\centering
\includegraphics[height=8cm]{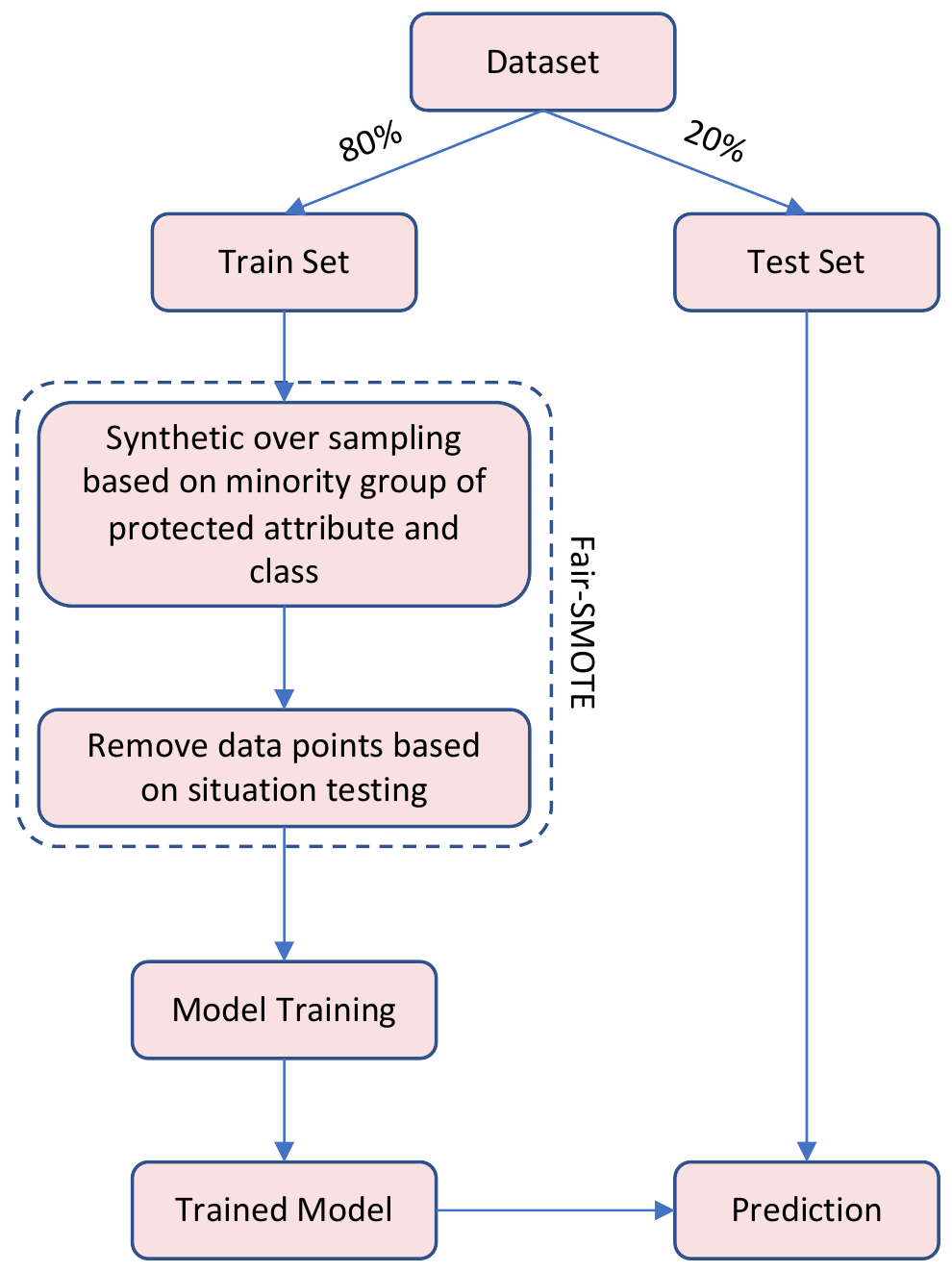}
\caption{Block diagram of Fair-SMOTE}
\label{block_diagram} 
\end{figure}

\subsection{Fair Situation Testing}
\label{situation_testing_subsetion}
At first, we use Fair-SMOTE to balance the training data. Fair-SMOTE is an oversampling technique and thus increases the size of train set. We then use \textit{situation testing} as mentioned in \S\ref{subsection:improper_label} to find out biased data points in the training data. We call this \textit{situation testing} as \textit{fair situation testing} because this is making the training data fairer. After finding biased data points, we remove them from training data. 
As Table \ref{Situation_Testing} shows small percentage numbers, we do not lose much of the training data. We will show in the ‘Results’ section that this
does not affect performance of the model much. After removal of biased data points, we train the model on the remaining training set and finally make the prediction on test data.

\subsection{Experimental Design}
Here we describe how we prepared the data for experiments to answer the research questions in \S\ref{results}. Our study used 10 datasets (Table \ref{Datasets}) and 3 classification models - logistic regression (LSR), random forest (RF), and support vector machine (SVM). In fairness domain, datasets are not very large in size and also have small dimensions. That is why we see most of the prior works ~\cite{Chakraborty_2020,10.1007/978-3-642-33486-3_3,NIPS2017_6988,Biswas_2020} choose simple models like us instead of deep learning models. For every experiment, we split the datasets using 5 fold cross-validation (train - 80\%, test - 20\%) and repeat 10 times with random seeds and finally report the median. The rows containing missing values are ignored, continuous features are converted to categorical (e.g., age<25: young, age>=25: old), non-numerical features are converted to numerical (e.g., male: 1, female: 0), finally, all the feature values are normalized (converted between 0 to 1). These basic conversions are done for all the experiments. Classification model is first trained on training data and then tested on test data; we report the median of ten runs.

\noindent
Figure \ref{block_diagram} shows the block-diagram of one repeat of our framework.

\subsection{Statistical Tests}
\label{subsetion_scott_knott}
 While comparing Fair-SMOTE with other techniques, we use Scott-Knott test \cite{7194626,6235961} to compare two distributions. Scott-Knott is a recursive bi-clustering algorithm that terminates when the difference between the two split groups is not significant. Scott-Knott searches for split points that
maximize the expected value of the difference between the means of the two resulting groups. If a group \textit{l} is split into two groups m and n, Scott-Knott searches for the split point that maximizes

\begin{align*}
    E[\Delta] = |m|/|l| (E[m] - E[l])^2 + |n|/|l| (E[n] - E[l])^2
\end{align*}

\noindent where $|m|$ is the size of group m. The result of the Scott-Knott test is ranks assigned to each result set; higher the rank, better the result. Scott-Knott
ranks two results the same if the difference between the distributions is not significant.

\section{Results}
\label{results}
The results are structured around five research questions.

\newenvironment{RQ}{\vspace{2mm}\begin{tcolorbox}[enhanced,width=3.3in,size=fbox,colback=lightpink,drop shadow southeast,sharp corners]}{\end{tcolorbox}}

\begin{RQ}
{\bf RQ1.} Can we find bias by just looking at the training data?
\end{RQ}
In \S\ref{root_cause} we said a machine learning model acquires bias from training data. RQ1 asks for the signals we must see to identify whether training data has bias or not. That is an important question to ask because if that is doable and that bias can be removed before model training, then the chances of bias affecting the final decision reduce significantly. Table \ref{RQ2_3} shows results for three different models (logistic regression (LSR), random forest (RF), support vector machine (SVM)), and Table \ref{RQ4_results} shows results for one model (LSR) only. The row ``Default'' is when we train the model on raw data. Results show that ``Default'' row has significantly high bias scores for all the datasets that means trained model is showing discrimination. Previously we have dived deep into these datasets to find reasons for bias. Here we are summarizing the sanity checks every dataset should go through before model training to avoid discrimination.

\begin{table*}[!t]
\caption{Results for RQ1, RQ2 \& RQ3.
In this table ``Default'' means off-the-shelf learner; SMOTE is an algorithm by Chawla et al.~\cite{Chawla_2002} from 2002;
and Fair-SMOTE is the algorithm introduced by this paper.
Cells show medians for 10 runs.
Here, the \colorbox{deeppink}{darkest} cells show top rank (note: for the metrics with `+' more is better and for the metrics with `-' less is better). The \colorbox{pink}{lighter} and \colorbox{lightpink}{lightest} cells show rank two and rank three respectively; the white cells show the worst rank. Rankings were calculated via Scott-Knott test (\S\ref{subsetion_scott_knott}).}
\label{RQ2_3}
\adjustbox{max width=5in}
{\begin{tabular}{|c|c|c|c|c|c|
>{\columncolor[HTML]{D28986}}c |
>{\columncolor[HTML]{D28986}}c |c|c|c|c|}
\hline
\cellcolor[HTML]{C0C0C0}Dataset                                                  & \cellcolor[HTML]{C0C0C0}\begin{tabular}[c]{@{}c@{}}Protected\\ Attribute\end{tabular} & \cellcolor[HTML]{C0C0C0}Algorithms & \cellcolor[HTML]{C0C0C0}\begin{tabular}[c]{@{}c@{}}Recall\\ (+)\end{tabular} & \cellcolor[HTML]{C0C0C0}\begin{tabular}[c]{@{}c@{}}False\\ alarm\\ (-)\end{tabular} & \cellcolor[HTML]{C0C0C0}\begin{tabular}[c]{@{}c@{}}Precision\\ (+)\end{tabular} & \cellcolor[HTML]{C0C0C0}\begin{tabular}[c]{@{}c@{}}Accuracy\\ (+)\end{tabular} & \cellcolor[HTML]{C0C0C0}\begin{tabular}[c]{@{}c@{}}F1 Score\\ (+)\end{tabular} & \cellcolor[HTML]{C0C0C0}\begin{tabular}[c]{@{}c@{}}AOD\\ (-)\end{tabular} & \cellcolor[HTML]{C0C0C0}\begin{tabular}[c]{@{}c@{}}EOD\\ (-)\end{tabular} & \cellcolor[HTML]{C0C0C0}\begin{tabular}[c]{@{}c@{}}SPD\\ (-)\end{tabular} & \cellcolor[HTML]{C0C0C0}\begin{tabular}[c]{@{}c@{}}DI\\ (-)\end{tabular} \\ \hline
                                                                                 &                                                                                       & Default - LSR                      & \cellcolor[HTML]{FFCCC9}{\color[HTML]{000000} 0.42}                          & \cellcolor[HTML]{D28986}{\color[HTML]{000000} 0.07}                                 & \cellcolor[HTML]{D28986}{\color[HTML]{000000} 0.69}                             & {\color[HTML]{000000} 0.83}                                                    & \cellcolor[HTML]{FFCCC9}{\color[HTML]{000000} 0.54}                            & \cellcolor[HTML]{FFCCC9}{\color[HTML]{000000} 0.12}                       & \cellcolor[HTML]{FFCCC9}{\color[HTML]{000000} 0.24}                       & \cellcolor[HTML]{FFCCC9}{\color[HTML]{000000} 0.21}                       & \cellcolor[HTML]{FFFFFF}{\color[HTML]{000000} 0.56}                      \\ \cline{3-12} 
                                                                                 &                                                                                       & Default - RF                       & \cellcolor[HTML]{FFCCC9}{\color[HTML]{000000} 0.51}                          & \cellcolor[HTML]{D28986}{\color[HTML]{000000} 0.06}                                 & \cellcolor[HTML]{D28986}{\color[HTML]{000000} 0.72}                             & {\color[HTML]{000000} 0.83}                                                    &  \cellcolor[HTML]{FFABA7}{\color[HTML]{000000} 0.59}                            &  \cellcolor[HTML]{FFABA7}{\color[HTML]{000000} 0.09}                       &  \cellcolor[HTML]{FFABA7}{\color[HTML]{000000} 0.17}                       & \cellcolor[HTML]{FFCCC9}{\color[HTML]{000000} 0.17}                       &  \cellcolor[HTML]{FFABA7}{\color[HTML]{000000} 0.33}                      \\ \cline{3-12} 
                                                                                 &                                                                                       & Default - SVM                      & \cellcolor[HTML]{FFFFFF}{\color[HTML]{000000} 0.35}                          & \cellcolor[HTML]{D28986}{\color[HTML]{000000} 0.04}                                 & \cellcolor[HTML]{D28986}{\color[HTML]{000000} 0.81}                             & {\color[HTML]{000000} 0.82}                                                    & \cellcolor[HTML]{FFFFFF}{\color[HTML]{000000} 0.49}                            & \cellcolor[HTML]{FFCCC9}{\color[HTML]{000000} 0.1}                        & \cellcolor[HTML]{FFCCC9}{\color[HTML]{000000} 0.24}                       &  \cellcolor[HTML]{FFABA7}{\color[HTML]{000000} 0.14}                       & \cellcolor[HTML]{FFCCC9}{\color[HTML]{000000} 0.43}                      \\ \cline{3-12} 
                                                                                 &                                                                                       & SMOTE - LSR                        & \cellcolor[HTML]{D28986}{\color[HTML]{000000} 0.70}                          & \cellcolor[HTML]{FFFFFF}{\color[HTML]{000000} 0.25}                                 & \cellcolor[HTML]{FFFFFF}{\color[HTML]{000000} 0.49}                             & \cellcolor[HTML]{FFFFFF}{\color[HTML]{000000} 0.70}                            & {\color[HTML]{000000} 0.64}                                                    & \cellcolor[HTML]{FFFFFF}{\color[HTML]{000000} 0.17}                       & \cellcolor[HTML]{FFFFFF}{\color[HTML]{000000} 0.37}                       & \cellcolor[HTML]{FFFFFF}{\color[HTML]{000000} 0.33}                       & \cellcolor[HTML]{FFFFFF}{\color[HTML]{000000} 0.58}                      \\ \cline{3-12} 
                                                                                 &                                                                                       & SMOTE - RF                         & \cellcolor[HTML]{FFCCC9}{\color[HTML]{000000} 0.61}                          & \cellcolor[HTML]{D28986}{\color[HTML]{000000} 0.07}                                 & \cellcolor[HTML]{D28986}{\color[HTML]{000000} 0.71}                             & {\color[HTML]{000000} 0.83}                                                    & {\color[HTML]{000000} 0.62}                                                    &  \cellcolor[HTML]{FFABA7}{\color[HTML]{000000} 0.08}                       &  \cellcolor[HTML]{FFABA7}{\color[HTML]{000000} 0.16}                       &  \cellcolor[HTML]{FFABA7}{\color[HTML]{000000} 0.16}                       &  \cellcolor[HTML]{FFABA7}{\color[HTML]{000000} 0.32}                      \\ \cline{3-12} 
                                                                                 &                                                                                       & SMOTE - SVM                        & \cellcolor[HTML]{D28986}{\color[HTML]{000000} 0.71}                          &  \cellcolor[HTML]{FFABA7}{\color[HTML]{000000} 0.19}                                 & \cellcolor[HTML]{FFCCC9}{\color[HTML]{000000} 0.55}                             &  \cellcolor[HTML]{FFABA7}{\color[HTML]{000000} 0.78}                            & {\color[HTML]{000000} 0.62}                                                    & \cellcolor[HTML]{FFCCC9}{\color[HTML]{000000} 0.11}                       & \cellcolor[HTML]{FFFFFF}{\color[HTML]{000000} 0.52}                       & \cellcolor[HTML]{FFFFFF}{\color[HTML]{000000} 0.42}                       & \cellcolor[HTML]{FFFFFF}{\color[HTML]{000000} 0.46}                      \\ \cline{3-12} 
                                                                                 &                                                                                       & Fair-SMOTE - LSR                   & \cellcolor[HTML]{D28986}{\color[HTML]{000000} 0.71}                          & \cellcolor[HTML]{FFFFFF}{\color[HTML]{000000} 0.25}                                 & \cellcolor[HTML]{FFFFFF}{\color[HTML]{000000} 0.51}                             & \cellcolor[HTML]{FFCCC9}{\color[HTML]{000000} 0.73}                            & {\color[HTML]{000000} 0.62}                                                    & \cellcolor[HTML]{D28986}{\color[HTML]{000000} 0.01}                       & \cellcolor[HTML]{D28986}{\color[HTML]{000000} 0.02}                       & \cellcolor[HTML]{D28986}{\color[HTML]{000000} 0.03}                       & \cellcolor[HTML]{D28986}{\color[HTML]{000000} 0.15}                      \\ \cline{3-12} 
                                                                                 &                                                                                       & Fair-SMOTE - RF                    & \cellcolor[HTML]{D28986}{\color[HTML]{000000} 0.69}                          &  \cellcolor[HTML]{FFABA7}{\color[HTML]{000000} 0.2}                                  & \cellcolor[HTML]{FFCCC9}{\color[HTML]{000000} 0.53}                             &  \cellcolor[HTML]{FFABA7}{\color[HTML]{000000} 0.78}                            & {\color[HTML]{000000} 0.6}                                                     & \cellcolor[HTML]{D28986}{\color[HTML]{000000} 0.03}                       & \cellcolor[HTML]{D28986}{\color[HTML]{000000} 0.04}                       & \cellcolor[HTML]{D28986}{\color[HTML]{000000} 0.1}                        & \cellcolor[HTML]{D28986}{\color[HTML]{000000} 0.22}                      \\ \cline{3-12} 
\multirow{-9}{*}{\begin{tabular}[c]{@{}c@{}}Adult Census \\ Income\end{tabular}} & \multirow{-9}{*}{Sex}                                                                 & Fair-SMOTE - SVM                   & \cellcolor[HTML]{D28986}{\color[HTML]{000000} 0.73}                          & \cellcolor[HTML]{FFFFFF}{\color[HTML]{000000} 0.23}                                 & \cellcolor[HTML]{FFFFFF}{\color[HTML]{000000} 0.51}                             &  \cellcolor[HTML]{FFABA7}{\color[HTML]{000000} 0.76}                            & {\color[HTML]{000000} 0.6}                                                     & \cellcolor[HTML]{D28986}{\color[HTML]{000000} 0.02}                       & \cellcolor[HTML]{D28986}{\color[HTML]{000000} 0.02}                       & \cellcolor[HTML]{D28986}{\color[HTML]{000000} 0.08}                       & \cellcolor[HTML]{D28986}{\color[HTML]{000000} 0.21}                      \\ \hline
                                                                                 &                                                                                       & Default - LSR                      & \cellcolor[HTML]{FFCCC9}{\color[HTML]{000000} 0.42}                          & \cellcolor[HTML]{D28986}{\color[HTML]{000000} 0.05}                                 & \cellcolor[HTML]{D28986}{\color[HTML]{000000} 0.69}                             & {\color[HTML]{000000} 0.81}                                                    &  \cellcolor[HTML]{FFABA7}{\color[HTML]{000000} 0.52}                            &  \cellcolor[HTML]{FFABA7}{\color[HTML]{000000} 0.06}                       & \cellcolor[HTML]{FFCCC9}{\color[HTML]{000000} 0.15}                       & \cellcolor[HTML]{FFCCC9}{\color[HTML]{000000} 0.16}                       & \cellcolor[HTML]{FFFFFF}{\color[HTML]{000000} 0.52}                      \\ \cline{3-12} 
                                                                                 &                                                                                       & Default - RF                       & \cellcolor[HTML]{FFCCC9}{\color[HTML]{000000} 0.53}                          & \cellcolor[HTML]{D28986}{\color[HTML]{000000} 0.07}                                 & \cellcolor[HTML]{D28986}{\color[HTML]{000000} 0.7}                              & {\color[HTML]{000000} 0.83}                                                    & {\color[HTML]{000000} 0.6}                                                     & \cellcolor[HTML]{FFCCC9}{\color[HTML]{000000} 0.12}                       & \cellcolor[HTML]{FFCCC9}{\color[HTML]{000000} 0.16}                       &  \cellcolor[HTML]{FFABA7}{\color[HTML]{000000} 0.12}                       & \cellcolor[HTML]{FFFFFF}{\color[HTML]{000000} 0.57}                      \\ \cline{3-12} 
                                                                                 &                                                                                       & Default - SVM                      & \cellcolor[HTML]{FFFFFF}{\color[HTML]{000000} 0.35}                          & \cellcolor[HTML]{D28986}{\color[HTML]{000000} 0.03}                                 & \cellcolor[HTML]{D28986}{\color[HTML]{000000} 0.8}                              & {\color[HTML]{000000} 0.82}                                                    &  \cellcolor[HTML]{FFABA7}{\color[HTML]{000000} 0.49}                            & \cellcolor[HTML]{FFCCC9}{\color[HTML]{000000} 0.08}                       &  \cellcolor[HTML]{FFABA7}{\color[HTML]{000000} 0.11}                       &  \cellcolor[HTML]{FFABA7}{\color[HTML]{000000} 0.10}                       &  \cellcolor[HTML]{FFABA7}{\color[HTML]{000000} 0.35}                      \\ \cline{3-12} 
                                                                                 &                                                                                       & SMOTE - LSR                        & \cellcolor[HTML]{D28986}{\color[HTML]{000000} 0.71}                          & \cellcolor[HTML]{FFFFFF}{\color[HTML]{000000} 0.23}                                 & \cellcolor[HTML]{FFFFFF}{\color[HTML]{000000} 0.49}                             &  \cellcolor[HTML]{FFABA7}{\color[HTML]{000000} 0.72}                            & {\color[HTML]{000000} 0.61}                                                    & \cellcolor[HTML]{FFFFFF}{\color[HTML]{000000} 0.16}                       & \cellcolor[HTML]{FFFFFF}{\color[HTML]{000000} 0.19}                       & \cellcolor[HTML]{FFFFFF}{\color[HTML]{000000} 0.23}                       & \cellcolor[HTML]{FFFFFF}{\color[HTML]{000000} 0.56}                      \\ \cline{3-12} 
                                                                                 &                                                                                       & SMOTE - RF                         &  \cellcolor[HTML]{FFABA7}{\color[HTML]{000000} 0.66}                          &  \cellcolor[HTML]{FFABA7}{\color[HTML]{000000} 0.08}                                 &  \cellcolor[HTML]{FFABA7}{\color[HTML]{000000} 0.67}                             & {\color[HTML]{000000} 0.83}                                                    & {\color[HTML]{000000} 0.63}                                                    & \cellcolor[HTML]{FFCCC9}{\color[HTML]{000000} 0.09}                       &  \cellcolor[HTML]{FFABA7}{\color[HTML]{000000} 0.12}                       &  \cellcolor[HTML]{FFABA7}{\color[HTML]{000000} 0.12}                       & \cellcolor[HTML]{FFFFFF}{\color[HTML]{000000} 0.52}                      \\ \cline{3-12} 
                                                                                 &                                                                                       & SMOTE - SVM                        & \cellcolor[HTML]{FFCCC9}{\color[HTML]{000000} 0.58}                          &  \cellcolor[HTML]{FFABA7}{\color[HTML]{000000} 0.11}                                 &  \cellcolor[HTML]{FFABA7}{\color[HTML]{000000} 0.62}                             & {\color[HTML]{000000} 0.81}                                                    & {\color[HTML]{000000} 0.6}                                                     & \cellcolor[HTML]{FFCCC9}{\color[HTML]{000000} 0.09}                       &  \cellcolor[HTML]{FFABA7}{\color[HTML]{000000} 0.12}                       & \cellcolor[HTML]{FFCCC9}{\color[HTML]{000000} 0.17}                       & \cellcolor[HTML]{FFFFFF}{\color[HTML]{000000} 0.32}                      \\ \cline{3-12} 
                                                                                 &                                                                                       & Fair-SMOTE - LSR                   & \cellcolor[HTML]{D28986}{\color[HTML]{000000} 0.7}                           & \cellcolor[HTML]{FFFFFF}{\color[HTML]{000000} 0.22}                                 & \cellcolor[HTML]{FFFFFF}{\color[HTML]{000000} 0.51}                             &  \cellcolor[HTML]{FFABA7}{\color[HTML]{000000} 0.72}                            & {\color[HTML]{000000} 0.62}                                                    & \cellcolor[HTML]{D28986}{\color[HTML]{000000} 0.04}                       & \cellcolor[HTML]{D28986}{\color[HTML]{000000} 0.03}                       & \cellcolor[HTML]{D28986}{\color[HTML]{000000} 0.05}                       & \cellcolor[HTML]{D28986}{\color[HTML]{000000} 0.26}                      \\ \cline{3-12} 
                                                                                 &                                                                                       & Fair-SMOTE - RF                    & \cellcolor[HTML]{D28986}{\color[HTML]{000000} 0.73}                          & \cellcolor[HTML]{FFCCC9}{\color[HTML]{000000} 0.2}                                  & \cellcolor[HTML]{FFCCC9}{\color[HTML]{000000} 0.52}                             & {\color[HTML]{000000} 0.8}                                                     & {\color[HTML]{000000} 0.61}                                                    & \cellcolor[HTML]{D28986}{\color[HTML]{000000} 0.01}                       & \cellcolor[HTML]{D28986}{\color[HTML]{000000} 0.02}                       & \cellcolor[HTML]{D28986}{\color[HTML]{000000} 0.08}                       & \cellcolor[HTML]{D28986}{\color[HTML]{000000} 0.29}                      \\ \cline{3-12} 
\multirow{-9}{*}{\begin{tabular}[c]{@{}c@{}}Adult Census \\ Income\end{tabular}} & \multirow{-9}{*}{Race}                                                                & Fair-SMOTE - SVM                   & \cellcolor[HTML]{D28986}{\color[HTML]{000000} 0.71}                          & \cellcolor[HTML]{FFFFFF}{\color[HTML]{000000} 0.26}                                 & \cellcolor[HTML]{FFFFFF}{\color[HTML]{000000} 0.5}                              & {\color[HTML]{000000} 0.75}                                                    & {\color[HTML]{000000} 0.61}                                                    & \cellcolor[HTML]{D28986}{\color[HTML]{000000} 0.01}                       & \cellcolor[HTML]{D28986}{\color[HTML]{000000} 0.01}                       & \cellcolor[HTML]{D28986}{\color[HTML]{000000} 0.06}                       & \cellcolor[HTML]{D28986}{\color[HTML]{000000} 0.18}                      \\ \hline
                                                                                 &                                                                                       & Default - LSR                      & \cellcolor[HTML]{D28986}{\color[HTML]{000000} 0.73}                          & \cellcolor[HTML]{D28986}{\color[HTML]{000000} 0.38}                                 & \cellcolor[HTML]{D28986}{\color[HTML]{000000} 0.66}                             & {\color[HTML]{000000} 0.64}                                                    & \cellcolor[HTML]{FFFFFF}{\color[HTML]{000000} 0.61}                            &  \cellcolor[HTML]{FFABA7}{\color[HTML]{000000} 0.05}                       &  \cellcolor[HTML]{FFABA7}{\color[HTML]{000000} 0.14}                       &  \cellcolor[HTML]{FFABA7}{\color[HTML]{000000} 0.18}                       &  \cellcolor[HTML]{FFABA7}{\color[HTML]{000000} 0.24}                      \\ \cline{3-12} 
                                                                                 &                                                                                       & Default - RF                       & \cellcolor[HTML]{D28986}{\color[HTML]{000000} 0.75}                          &  \cellcolor[HTML]{FFABA7}{\color[HTML]{000000} 0.45}                                 & \cellcolor[HTML]{D28986}{\color[HTML]{000000} 0.66}                             & {\color[HTML]{000000} 0.67}                                                    & {\color[HTML]{000000} 0.72}                                                    & \cellcolor[HTML]{FFFFFF}{\color[HTML]{000000} 0.11}                       & \cellcolor[HTML]{FFFFFF}{\color[HTML]{000000} 0.18}                       &  \cellcolor[HTML]{FFABA7}{\color[HTML]{000000} 0.2}                        & \cellcolor[HTML]{FFFFFF}{\color[HTML]{000000} 0.34}                      \\ \cline{3-12} 
                                                                                 &                                                                                       & Default - SVM                      & \cellcolor[HTML]{D28986}{\color[HTML]{000000} 0.77}                          &  \cellcolor[HTML]{FFABA7}{\color[HTML]{000000} 0.45}                                 & \cellcolor[HTML]{D28986}{\color[HTML]{000000} 0.66}                             & {\color[HTML]{000000} 0.67}                                                    & {\color[HTML]{000000} 0.71}                                                    & \cellcolor[HTML]{FFFFFF}{\color[HTML]{000000} 0.1}                        &  \cellcolor[HTML]{FFABA7}{\color[HTML]{000000} 0.15}                       & \cellcolor[HTML]{FFFFFF}{\color[HTML]{000000} 0.22}                       & \cellcolor[HTML]{FFFFFF}{\color[HTML]{000000} 0.28}                      \\ \cline{3-12} 
                                                                                 &                                                                                       & SMOTE - LSR                        & \cellcolor[HTML]{FFCCC9}{\color[HTML]{000000} 0.65}                          & \cellcolor[HTML]{D28986}{\color[HTML]{000000} 0.33}                                 &  \cellcolor[HTML]{FFABA7}{\color[HTML]{000000} 0.62}                             &  \cellcolor[HTML]{FFABA7}{\color[HTML]{000000} 0.6}                             &  \cellcolor[HTML]{FFABA7}{\color[HTML]{000000} 0.65}                            & \cellcolor[HTML]{FFFFFF}{\color[HTML]{000000} 0.08}                       & \cellcolor[HTML]{FFFFFF}{\color[HTML]{000000} 0.19}                       & \cellcolor[HTML]{FFFFFF}{\color[HTML]{000000} 0.22}                       & \cellcolor[HTML]{FFFFFF}{\color[HTML]{000000} 0.31}                      \\ \cline{3-12} 
                                                                                 &                                                                                       & SMOTE - RF                         &  \cellcolor[HTML]{FFABA7}{\color[HTML]{000000} 0.72}                          &  \cellcolor[HTML]{FFABA7}{\color[HTML]{000000} 0.42}                                 & \cellcolor[HTML]{D28986}{\color[HTML]{000000} 0.67}                             & {\color[HTML]{000000} 0.65}                                                    &  \cellcolor[HTML]{FFABA7}{\color[HTML]{000000} 0.7}                             & \cellcolor[HTML]{FFFFFF}{\color[HTML]{000000} 0.11}                       & \cellcolor[HTML]{FFFFFF}{\color[HTML]{000000} 0.22}                       & \cellcolor[HTML]{FFFFFF}{\color[HTML]{000000} 0.26}                       & \cellcolor[HTML]{FFFFFF}{\color[HTML]{000000} 0.3}                       \\ \cline{3-12} 
                                                                                 &                                                                                       & SMOTE - SVM                        &  \cellcolor[HTML]{FFABA7}{\color[HTML]{000000} 0.7}                           & \cellcolor[HTML]{D28986}{\color[HTML]{000000} 0.36}                                 & \cellcolor[HTML]{D28986}{\color[HTML]{000000} 0.68}                             & {\color[HTML]{000000} 0.66}                                                    & {\color[HTML]{000000} 0.69}                                                    & \cellcolor[HTML]{FFFFFF}{\color[HTML]{000000} 0.1}                        & \cellcolor[HTML]{FFFFFF}{\color[HTML]{000000} 0.21}                       & \cellcolor[HTML]{FFFFFF}{\color[HTML]{000000} 0.31}                       & \cellcolor[HTML]{FFFFFF}{\color[HTML]{000000} 0.38}                      \\ \cline{3-12} 
                                                                                 &                                                                                       & Fair-SMOTE - LSR                   & \cellcolor[HTML]{FFFFFF}{\color[HTML]{000000} 0.62}                          & \cellcolor[HTML]{D28986}{\color[HTML]{000000} 0.32}                                 & \cellcolor[HTML]{FFFFFF}{\color[HTML]{000000} 0.56}                             &  \cellcolor[HTML]{FFABA7}{\color[HTML]{000000} 0.55}                            &  \cellcolor[HTML]{FFABA7}{\color[HTML]{000000} 0.65}                            & \cellcolor[HTML]{D28986}{\color[HTML]{000000} 0.02}                       & \cellcolor[HTML]{D28986}{\color[HTML]{000000} 0.05}                       & \cellcolor[HTML]{D28986}{\color[HTML]{000000} 0.08}                       & \cellcolor[HTML]{D28986}{\color[HTML]{000000} 0.04}                      \\ \cline{3-12} 
                                                                                 &                                                                                       & Fair-SMOTE - RF                    &  \cellcolor[HTML]{FFABA7}{\color[HTML]{000000} 0.71}                          &  \cellcolor[HTML]{FFABA7}{\color[HTML]{000000} 0.44}                                 & \cellcolor[HTML]{D28986}{\color[HTML]{000000} 0.66}                             & {\color[HTML]{000000} 0.65}                                                    & {\color[HTML]{000000} 0.7}                                                     & \cellcolor[HTML]{D28986}{\color[HTML]{000000} 0.04}                       & \cellcolor[HTML]{D28986}{\color[HTML]{000000} 0.03}                       & \cellcolor[HTML]{D28986}{\color[HTML]{000000} 0.1}                        & \cellcolor[HTML]{D28986}{\color[HTML]{000000} 0.02}                      \\ \cline{3-12} 
\multirow{-9}{*}{Compas}                                                         & \multirow{-9}{*}{Sex}                                                                 & Fair-SMOTE - SVM                   & \cellcolor[HTML]{D28986}{\color[HTML]{000000} 0.79}                          & \cellcolor[HTML]{FFFFFF}{\color[HTML]{000000} 0.5}                                  & \cellcolor[HTML]{D28986}{\color[HTML]{000000} 0.65}                             & {\color[HTML]{000000} 0.66}                                                    & {\color[HTML]{000000} 0.71}                                                    & \cellcolor[HTML]{D28986}{\color[HTML]{000000} 0.02}                       & \cellcolor[HTML]{D28986}{\color[HTML]{000000} 0.01}                       & \cellcolor[HTML]{D28986}{\color[HTML]{000000} 0.06}                       & \cellcolor[HTML]{D28986}{\color[HTML]{000000} 0.08}                      \\ \hline
                                                                                 &                                                                                       & Default - LSR                      & \cellcolor[HTML]{D28986}{\color[HTML]{000000} 0.69}                          & \cellcolor[HTML]{FFFFFF}{\color[HTML]{000000} 0.39}                                 & \cellcolor[HTML]{D28986}{\color[HTML]{000000} 0.65}                             & {\color[HTML]{000000} 0.64}                                                    &  \cellcolor[HTML]{FFABA7}{\color[HTML]{000000} 0.68}                            &  \cellcolor[HTML]{FFABA7}{\color[HTML]{000000} 0.05}                       &  \cellcolor[HTML]{FFABA7}{\color[HTML]{000000} 0.11}                       &  \cellcolor[HTML]{FFABA7}{\color[HTML]{000000} 0.12}                       &  \cellcolor[HTML]{FFABA7}{\color[HTML]{000000} 0.21}                      \\ \cline{3-12} 
                                                                                 &                                                                                       & Default - RF                       & \cellcolor[HTML]{D28986}{\color[HTML]{000000} 0.75}                          & \cellcolor[HTML]{FFFFFF}{\color[HTML]{000000} 0.44}                                 & \cellcolor[HTML]{D28986}{\color[HTML]{000000} 0.66}                             & {\color[HTML]{000000} 0.66}                                                    & {\color[HTML]{000000} 0.7}                                                     & \cellcolor[HTML]{FFFFFF}{\color[HTML]{000000} 0.07}                       & \cellcolor[HTML]{FFFFFF}{\color[HTML]{000000} 0.17}                       & \cellcolor[HTML]{FFFFFF}{\color[HTML]{000000} 0.21}                       &  \cellcolor[HTML]{FFABA7}{\color[HTML]{000000} 0.24}                      \\ \cline{3-12} 
                                                                                 &                                                                                       & Default - SVM                      & \cellcolor[HTML]{D28986}{\color[HTML]{000000} 0.77}                          & \cellcolor[HTML]{FFFFFF}{\color[HTML]{000000} 0.45}                                 & \cellcolor[HTML]{D28986}{\color[HTML]{000000} 0.66}                             & {\color[HTML]{000000} 0.67}                                                    & {\color[HTML]{000000} 0.71}                                                    & \cellcolor[HTML]{FFFFFF}{\color[HTML]{000000} 0.07}                       & \cellcolor[HTML]{FFCCC9}{\color[HTML]{000000} 0.14}                       & \cellcolor[HTML]{FFFFFF}{\color[HTML]{000000} 0.18}                       &  \cellcolor[HTML]{FFABA7}{\color[HTML]{000000} 0.24}                      \\ \cline{3-12} 
                                                                                 &                                                                                       & SMOTE - LSR                        &  \cellcolor[HTML]{FFABA7}{\color[HTML]{000000} 0.61}                          & \cellcolor[HTML]{D28986}{\color[HTML]{000000} 0.32}                                 &  \cellcolor[HTML]{FFABA7}{\color[HTML]{000000} 0.61}                             &  \cellcolor[HTML]{FFABA7}{\color[HTML]{000000} 0.6}                             &  \cellcolor[HTML]{FFABA7}{\color[HTML]{000000} 0.63}                            & \cellcolor[HTML]{FFFFFF}{\color[HTML]{000000} 0.06}                       & \cellcolor[HTML]{FFFFFF}{\color[HTML]{000000} 0.16}                       &  \cellcolor[HTML]{FFABA7}{\color[HTML]{000000} 0.14}                       & \cellcolor[HTML]{FFFFFF}{\color[HTML]{000000} 0.27}                      \\ \cline{3-12} 
                                                                                 &                                                                                       & SMOTE - RF                         & \cellcolor[HTML]{D28986}{\color[HTML]{000000} 0.75}                          &  \cellcolor[HTML]{FFABA7}{\color[HTML]{000000} 0.42}                                 & \cellcolor[HTML]{D28986}{\color[HTML]{000000} 0.67}                             & {\color[HTML]{000000} 0.66}                                                    & {\color[HTML]{000000} 0.7}                                                     & \cellcolor[HTML]{FFFFFF}{\color[HTML]{000000} 0.07}                       &  \cellcolor[HTML]{FFABA7}{\color[HTML]{000000} 0.13}                       & \cellcolor[HTML]{FFFFFF}{\color[HTML]{000000} 0.19}                       & \cellcolor[HTML]{FFFFFF}{\color[HTML]{000000} 0.31}                      \\ \cline{3-12} 
                                                                                 &                                                                                       & SMOTE - SVM                        & \cellcolor[HTML]{D28986}{\color[HTML]{000000} 0.7}                           &  \cellcolor[HTML]{FFABA7}{\color[HTML]{000000} 0.39}                                 & \cellcolor[HTML]{D28986}{\color[HTML]{000000} 0.68}                             & {\color[HTML]{000000} 0.66}                                                    & {\color[HTML]{000000} 0.69}                                                    & \cellcolor[HTML]{FFFFFF}{\color[HTML]{000000} 0.09}                       &  \cellcolor[HTML]{FFABA7}{\color[HTML]{000000} 0.12}                       & \cellcolor[HTML]{FFFFFF}{\color[HTML]{000000} 0.16}                       &  \cellcolor[HTML]{FFABA7}{\color[HTML]{000000} 0.24}                      \\ \cline{3-12} 
                                                                                 &                                                                                       & Fair-SMOTE - LSR                   &  \cellcolor[HTML]{FFABA7}{\color[HTML]{000000} 0.62}                          & \cellcolor[HTML]{D28986}{\color[HTML]{000000} 0.30}                                 &  \cellcolor[HTML]{FFABA7}{\color[HTML]{000000} 0.56}                             & {\color[HTML]{000000} 0.55}                                                    & {\color[HTML]{000000} 0.66}                                                    & \cellcolor[HTML]{D28986}{\color[HTML]{000000} 0.01}                       & \cellcolor[HTML]{D28986}{\color[HTML]{000000} 0.05}                       & \cellcolor[HTML]{D28986}{\color[HTML]{000000} 0.06}                       & \cellcolor[HTML]{D28986}{\color[HTML]{000000} 0.11}                      \\ \cline{3-12} 
                                                                                 &                                                                                       & Fair-SMOTE - RF                    &  \cellcolor[HTML]{FFABA7}{\color[HTML]{000000} 0.66}                          &  \cellcolor[HTML]{FFABA7}{\color[HTML]{000000} 0.39}                                 & \cellcolor[HTML]{D28986}{\color[HTML]{000000} 0.67}                             & {\color[HTML]{000000} 0.65}                                                    & {\color[HTML]{000000} 0.67}                                                    & \cellcolor[HTML]{D28986}{\color[HTML]{000000} 0.01}                       & \cellcolor[HTML]{D28986}{\color[HTML]{000000} 0.03}                       & \cellcolor[HTML]{D28986}{\color[HTML]{000000} 0.02}                       & \cellcolor[HTML]{D28986}{\color[HTML]{000000} 0.10}                      \\ \cline{3-12} 
\multirow{-9}{*}{Compas}                                                         & \multirow{-9}{*}{Race}                                                                & Fair-SMOTE - SVM                   & \cellcolor[HTML]{D28986}{\color[HTML]{000000} 0.7}                           &  \cellcolor[HTML]{FFABA7}{\color[HTML]{000000} 0.41}                                 & \cellcolor[HTML]{D28986}{\color[HTML]{000000} 0.67}                             & {\color[HTML]{000000} 0.65}                                                    & {\color[HTML]{000000} 0.68}                                                    & \cellcolor[HTML]{D28986}{\color[HTML]{000000} 0.02}                       & \cellcolor[HTML]{D28986}{\color[HTML]{000000} 0.06}                       & \cellcolor[HTML]{D28986}{\color[HTML]{000000} 0.08}                       & \cellcolor[HTML]{D28986}{\color[HTML]{000000} 0.12}                      \\ \hline
                                                                                 &                                                                                       & Default - LSR                      &  \cellcolor[HTML]{FFABA7}{\color[HTML]{000000} 0.35}                          &  \cellcolor[HTML]{D28986}{\color[HTML]{000000} 0.05}                                 &  \cellcolor[HTML]{D28986}{\color[HTML]{000000} 0.65}                             &  \cellcolor[HTML]{D28986}{\color[HTML]{000000} 0.85}                            &  \cellcolor[HTML]{FFABA7}{\color[HTML]{000000} 0.44}                            &  \cellcolor[HTML]{FFABA7}{\color[HTML]{000000} 0.04}                       &  \cellcolor[HTML]{FFABA7}{\color[HTML]{000000} 0.1}                        &  \cellcolor[HTML]{FFABA7}{\color[HTML]{000000} 0.07}                       & \cellcolor[HTML]{FFCCC9}{\color[HTML]{000000} 0.43}                      \\ \cline{3-12} 
                                                                                 &                                                                                       & Default - RF                       &  \cellcolor[HTML]{FFABA7}{\color[HTML]{000000} 0.38}                          &  \cellcolor[HTML]{FFABA7}{\color[HTML]{000000} 0.11}                                 &  \cellcolor[HTML]{FFABA7}{\color[HTML]{000000} 0.62}                             &  \cellcolor[HTML]{FFABA7}{\color[HTML]{000000} 0.81}                            &  \cellcolor[HTML]{FFABA7}{\color[HTML]{000000} 0.41}                            & \cellcolor[HTML]{FFCCC9}{\color[HTML]{000000} 0.08}                       &  \cellcolor[HTML]{FFABA7}{\color[HTML]{000000} 0.12}                       &  \cellcolor[HTML]{FFABA7}{\color[HTML]{000000} 0.08}                       &  \cellcolor[HTML]{FFABA7}{\color[HTML]{000000} 0.36}                      \\ \cline{3-12} 
                                                                                 &                                                                                       & Default - SVM                      &  \cellcolor[HTML]{FFABA7}{\color[HTML]{000000} 0.31}                          &  \cellcolor[HTML]{D28986}{\color[HTML]{000000} 0.08}                                 &  \cellcolor[HTML]{D28986}{\color[HTML]{000000} 0.66}                             &  \cellcolor[HTML]{FFABA7}{\color[HTML]{000000} 0.79}                            &  \cellcolor[HTML]{FFABA7}{\color[HTML]{000000} 0.4}                             & \cellcolor[HTML]{FFCCC9}{\color[HTML]{000000} 0.09}                       &  \cellcolor[HTML]{FFABA7}{\color[HTML]{000000} 0.1}                        &  \cellcolor[HTML]{FFABA7}{\color[HTML]{000000} 0.09}                       &  \cellcolor[HTML]{FFABA7}{\color[HTML]{000000} 0.32}                      \\ \cline{3-12} 
                                                                                 &                                                                                       & SMOTE - LSR                        &  \cellcolor[HTML]{D28986}{\color[HTML]{000000} 0.65}                          & \cellcolor[HTML]{FFCCC9}{\color[HTML]{000000} 0.22}                                 & \cellcolor[HTML]{FFCCC9}{\color[HTML]{000000} 0.58}                             &  \cellcolor[HTML]{FFABA7}{\color[HTML]{000000} 0.78}                            &  \cellcolor[HTML]{D28986}{\color[HTML]{000000} 0.49}                            & \cellcolor[HTML]{FFFFFF}{\color[HTML]{000000} 0.17}                       &  \cellcolor[HTML]{FFABA7}{\color[HTML]{000000} 0.11}                       & \cellcolor[HTML]{FFCCC9}{\color[HTML]{000000} 0.11}                       & \cellcolor[HTML]{FFCCC9}{\color[HTML]{000000} 0.49}                      \\ \cline{3-12} 
                                                                                 &                                                                                       & SMOTE - RF                         &  \cellcolor[HTML]{D28986}{\color[HTML]{000000} 0.63}                          & \cellcolor[HTML]{FFCCC9}{\color[HTML]{000000} 0.18}                                 & \cellcolor[HTML]{FFCCC9}{\color[HTML]{000000} 0.56}                             &  \cellcolor[HTML]{FFABA7}{\color[HTML]{000000} 0.78}                            &  \cellcolor[HTML]{D28986}{\color[HTML]{000000} 0.52}                            & \cellcolor[HTML]{FFFFFF}{\color[HTML]{000000} 0.12}                       & \cellcolor[HTML]{FFCCC9}{\color[HTML]{000000} 0.19}                       & \cellcolor[HTML]{FFCCC9}{\color[HTML]{000000} 0.13}                       &  \cellcolor[HTML]{FFABA7}{\color[HTML]{000000} 0.31}                      \\ \cline{3-12} 
                                                                                 &                                                                                       & SMOTE - SVM                        &  \cellcolor[HTML]{D28986}{\color[HTML]{000000} 0.62}                          & \cellcolor[HTML]{FFCCC9}{\color[HTML]{000000} 0.23}                                 & \cellcolor[HTML]{FFFFFF}{\color[HTML]{000000} 0.49}                             &  \cellcolor[HTML]{FFABA7}{\color[HTML]{000000} 0.79}                            &  \cellcolor[HTML]{D28986}{\color[HTML]{000000} 0.5}                             & \cellcolor[HTML]{FFFFFF}{\color[HTML]{000000} 0.13}                       & \cellcolor[HTML]{FFCCC9}{\color[HTML]{000000} 0.18}                       & \cellcolor[HTML]{FFCCC9}{\color[HTML]{000000} 0.15}                       & \cellcolor[HTML]{FFCCC9}{\color[HTML]{000000} 0.47}                      \\ \cline{3-12} 
                                                                                 &                                                                                       & Fair-SMOTE - LSR                   &  \cellcolor[HTML]{D28986}{\color[HTML]{000000} 0.66}                          & \cellcolor[HTML]{FFCCC9}{\color[HTML]{000000} 0.2}                                  & \cellcolor[HTML]{FFFFFF}{\color[HTML]{000000} 0.41}                             &  \cellcolor[HTML]{FFABA7}{\color[HTML]{000000} 0.77}                            &  \cellcolor[HTML]{D28986}{\color[HTML]{000000} 0.51}                            &  \cellcolor[HTML]{D28986}{\color[HTML]{000000} 0.01}                       &  \cellcolor[HTML]{D28986}{\color[HTML]{000000} 0.03}                       &  \cellcolor[HTML]{D28986}{\color[HTML]{000000} 0.04}                       &  \cellcolor[HTML]{D28986}{\color[HTML]{000000} 0.17}                      \\ \cline{3-12} 
                                                                                 &                                                                                       & Fair-SMOTE - RF                    &  \cellcolor[HTML]{D28986}{\color[HTML]{000000} 0.62}                          & \cellcolor[HTML]{FFCCC9}{\color[HTML]{000000} 0.21}                                 & \cellcolor[HTML]{FFFFFF}{\color[HTML]{000000} 0.38}                             &  \cellcolor[HTML]{FFABA7}{\color[HTML]{000000} 0.77}                            &  \cellcolor[HTML]{D28986}{\color[HTML]{000000} 0.48}                            &  \cellcolor[HTML]{D28986}{\color[HTML]{000000} 0.03}                       &  \cellcolor[HTML]{D28986}{\color[HTML]{000000} 0.04}                       &  \cellcolor[HTML]{D28986}{\color[HTML]{000000} 0.03}                       &  \cellcolor[HTML]{D28986}{\color[HTML]{000000} 0.19}                      \\ \cline{3-12} 
\multirow{-9}{*}{MEPS - 16}                                                      & \multirow{-9}{*}{Race}                                                                 & Fair-SMOTE - SVM                   &  \cellcolor[HTML]{D28986}{\color[HTML]{000000} 0.61}                          & \cellcolor[HTML]{FFCCC9}{\color[HTML]{000000} 0.18}                                 & \cellcolor[HTML]{FFFFFF}{\color[HTML]{000000} 0.39}                             & \cellcolor[HTML]{FFCCC9}{\color[HTML]{000000} 0.76}                            &  \cellcolor[HTML]{D28986}{\color[HTML]{000000} 0.49}                            &  \cellcolor[HTML]{D28986}{\color[HTML]{000000} 0.04}                       &  \cellcolor[HTML]{D28986}{\color[HTML]{000000} 0.03}                       &  \cellcolor[HTML]{D28986}{\color[HTML]{000000} 0.05}                       &  \cellcolor[HTML]{D28986}{\color[HTML]{000000} 0.19}                      \\ \hline
\end{tabular}}
\end{table*}

\begin{table*}[]
\caption{Results for RQ3, RQ4  (learner= Logistic Regression).
In this table. ``Default'' denotes off-the-shelf logistic regression; OP is Calmon et al.'s system from NIPS'17~\cite{NIPS2017_6988};
Fairway is 
Chakraborty et al.'s system from FSE'20~\cite{Chakraborty_2020};
and Fair-SMOTE is the  algorithm introduced by this paper.
Cells show medians for 10 runs. Here,  the \colorbox{deeppink}{darker} cells show top rank (note: for the metrics with `+' more is better and for the metrics with `-' less is better). The \colorbox{lightpink}{lighter} cells show rank two; white shows lowest rank (worst performance).
Rankings were calculated via the
Scott-Knott test (\S\ref{subsetion_scott_knott})}
\label{RQ4_results}
\adjustbox{max width=5in}{\begin{tabular}{|c|c|c|c|
>{\columncolor[HTML]{D28986}}c |
>{\columncolor[HTML]{D28986}}c |c|
>{\columncolor[HTML]{FFCCC9}}c |
>{\columncolor[HTML]{D28986}}c |
>{\columncolor[HTML]{D28986}}c |
>{\columncolor[HTML]{D28986}}c |
>{\columncolor[HTML]{D28986}}c |}
\hline
\cellcolor[HTML]{C0C0C0}{\color[HTML]{000000} Dataset}                                                  & \cellcolor[HTML]{C0C0C0}{\color[HTML]{000000} \begin{tabular}[c]{@{}c@{}}Protected\\ Attribute\end{tabular}} & \cellcolor[HTML]{C0C0C0}{\color[HTML]{000000} Algorithms} & \cellcolor[HTML]{C0C0C0}{\color[HTML]{000000} \begin{tabular}[c]{@{}c@{}}Recall\\ (+)\end{tabular}} & \cellcolor[HTML]{C0C0C0}{\color[HTML]{000000} \begin{tabular}[c]{@{}c@{}}False alarm\\ (-)\end{tabular}} & \cellcolor[HTML]{C0C0C0}{\color[HTML]{000000} \begin{tabular}[c]{@{}c@{}}Precision\\ (+)\end{tabular}} & \cellcolor[HTML]{C0C0C0}{\color[HTML]{000000} \begin{tabular}[c]{@{}c@{}}Accuracy\\ (+)\end{tabular}} & \cellcolor[HTML]{C0C0C0}{\color[HTML]{000000} \begin{tabular}[c]{@{}c@{}}F1 Score\\ (+)\end{tabular}} & \cellcolor[HTML]{C0C0C0}{\color[HTML]{000000} \begin{tabular}[c]{@{}c@{}}AOD\\ (-)\end{tabular}} & \cellcolor[HTML]{C0C0C0}{\color[HTML]{000000} \begin{tabular}[c]{@{}c@{}}EOD\\ (-)\end{tabular}} & \cellcolor[HTML]{C0C0C0}{\color[HTML]{000000} \begin{tabular}[c]{@{}c@{}}SPD\\ (-)\end{tabular}} & \cellcolor[HTML]{C0C0C0}{\color[HTML]{000000} \begin{tabular}[c]{@{}c@{}}DI\\ (-)\end{tabular}} \\ \hline
{\color[HTML]{000000} }                                                                                 & {\color[HTML]{000000} }                                                                                      & {\color[HTML]{000000} Default}                            & \cellcolor[HTML]{FFCCC9}{\color[HTML]{000000}  {0.42}}                                        & \cellcolor[HTML]{FFCCC9}{\color[HTML]{000000}  {0.07}}                                             & {\color[HTML]{000000}  {0.69}}                                                                   & \cellcolor[HTML]{D28986}{\color[HTML]{000000}  {0.83}}                                          & {\color[HTML]{000000}  {0.54}}                                                                  & \cellcolor[HTML]{FFFFFF}{\color[HTML]{000000}  {0.12}}                                     & \cellcolor[HTML]{FFFFFF}{\color[HTML]{000000}  {0.24}}                                     & \cellcolor[HTML]{FFFFFF}{\color[HTML]{000000}  {0.21}}                                     & \cellcolor[HTML]{FFFFFF}{\color[HTML]{000000}  {0.56}}                                    \\ \cline{3-12} 
{\color[HTML]{000000} }                                                                                 & {\color[HTML]{000000} }                                                                                      & {\color[HTML]{000000} OP}            & \cellcolor[HTML]{FFCCC9}{\color[HTML]{000000}  {0.41}}                                        & \cellcolor[HTML]{FFCCC9}{\color[HTML]{000000}  {0.09}}                                             & \cellcolor[HTML]{FFCCC9}{\color[HTML]{000000}  {0.61}}                                           & \cellcolor[HTML]{FFCCC9}{\color[HTML]{000000}  {0.76}}                                          & {\color[HTML]{000000}  {0.51}}                                                                  & {\color[HTML]{000000}  {0.04}}                                                             & {\color[HTML]{000000}  {0.03}}                                                             & {\color[HTML]{000000}  {0.04}}                                                             & {\color[HTML]{000000}  {0.14}}                                                            \\ \cline{3-12} 
{\color[HTML]{000000} }                                                                                 & {\color[HTML]{000000} }                                                                                      & {\color[HTML]{000000} Fairway}                            & \cellcolor[HTML]{FFFFFF}{\color[HTML]{000000}  {0.25}}                                        & {\color[HTML]{000000}  {0.04}}                                                                     & {\color[HTML]{000000}  {0.70}}                                                                   & \cellcolor[HTML]{FFCCC9}{\color[HTML]{000000}  {0.72}}                                          & \cellcolor[HTML]{FFFFFF}{\color[HTML]{000000}  {0.42}}                                          & {\color[HTML]{000000}  {0.02}}                                                             & {\color[HTML]{000000}  {0.03}}                                                             & {\color[HTML]{000000}  {0.01}}                                                             & {\color[HTML]{000000}  {0.11}}                                                            \\ \cline{3-12} 
\multirow{-4}{*}{{\color[HTML]{000000} \begin{tabular}[c]{@{}c@{}}Adult Census \\ Income\end{tabular}}} & \multirow{-4}{*}{{\color[HTML]{000000} Sex}}                                                                 & {\color[HTML]{000000} Fair SMOTE}                         & \cellcolor[HTML]{D28986}{\color[HTML]{000000}  {0.71}}                                        & \cellcolor[HTML]{FFFFFF}{\color[HTML]{000000}  {0.25}}                                             & \cellcolor[HTML]{FFFFFF}{\color[HTML]{000000}  {0.51}}                                           & \cellcolor[HTML]{FFCCC9}{\color[HTML]{000000}  {0.73}}                                          & \cellcolor[HTML]{D28986}{\color[HTML]{000000}  {0.62}}                                          & {\color[HTML]{000000}  {0.01}}                                                             & {\color[HTML]{000000}  {0.02}}                                                             & {\color[HTML]{000000}  {0.03}}                                                             & {\color[HTML]{000000}  {0.12}}                                                            \\ \hline
{\color[HTML]{000000} }                                                                                 & {\color[HTML]{000000} }                                                                                      & {\color[HTML]{000000} Default}                            & \cellcolor[HTML]{FFCCC9}{\color[HTML]{000000}  {0.42}}                                        & {\color[HTML]{000000}  {0.05}}                                                                     & {\color[HTML]{000000}  {0.69}}                                                                   & \cellcolor[HTML]{D28986}{\color[HTML]{000000}  {0.81}}                                          & {\color[HTML]{000000}  {0.52}}                                                                  & \cellcolor[HTML]{FFFFFF}{\color[HTML]{000000}  {0.06}}                                     & \cellcolor[HTML]{FFFFFF}{\color[HTML]{000000}  {0.15}}                                     & \cellcolor[HTML]{FFFFFF}{\color[HTML]{000000}  {0.16}}                                     & \cellcolor[HTML]{FFFFFF}{\color[HTML]{000000}  {0.52}}                                    \\ \cline{3-12} 
{\color[HTML]{000000} }                                                                                 & {\color[HTML]{000000} }                                                                                      & {\color[HTML]{000000} OP}            & \cellcolor[HTML]{FFFFFF}{\color[HTML]{000000}  {0.38}}                                        & {\color[HTML]{000000}  {0.06}}                                                                     & {\color[HTML]{000000}  {0.66}}                                                                   & \cellcolor[HTML]{D28986}{\color[HTML]{000000}  {0.78}}                                          & {\color[HTML]{000000}  {0.48}}                                                                  & {\color[HTML]{000000}  {0.03}}                                                             & {\color[HTML]{000000}  {0.02}}                                                             & {\color[HTML]{000000}  {0.05}}                                                             & {\color[HTML]{000000}  {0.21}}                                                            \\ \cline{3-12} 
{\color[HTML]{000000} }                                                                                 & {\color[HTML]{000000} }                                                                                      & {\color[HTML]{000000} Fairway}                            & \cellcolor[HTML]{FFFFFF}{\color[HTML]{000000}  {0.36}}                                        & {\color[HTML]{000000}  {0.04}}                                                                     & {\color[HTML]{000000}  {0.70}}                                                                   & \cellcolor[HTML]{FFCCC9}{\color[HTML]{000000}  {0.73}}                                          & \cellcolor[HTML]{FFFFFF}{\color[HTML]{000000}  {0.44}}                                          & {\color[HTML]{000000}  {0.02}}                                                             & {\color[HTML]{000000}  {0.03}}                                                             & {\color[HTML]{000000}  {0.06}}                                                             & \cellcolor[HTML]{FFCCC9}{\color[HTML]{000000}  {0.32}}                                    \\ \cline{3-12} 
\multirow{-4}{*}{{\color[HTML]{000000} \begin{tabular}[c]{@{}c@{}}Adult Census \\ Income\end{tabular}}} & \multirow{-4}{*}{{\color[HTML]{000000} Race}}                                                                & {\color[HTML]{000000} Fair SMOTE}                         & \cellcolor[HTML]{D28986}{\color[HTML]{000000}  {0.7}}                                         & \cellcolor[HTML]{FFCCC9}{\color[HTML]{000000}  {0.22}}                                             & \cellcolor[HTML]{FFCCC9}{\color[HTML]{000000}  {0.51}}                                           & \cellcolor[HTML]{FFCCC9}{\color[HTML]{000000}  {0.72}}                                          & \cellcolor[HTML]{D28986}{\color[HTML]{000000}  {0.62}}                                          & {\color[HTML]{000000}  {0.04}}                                                             & {\color[HTML]{000000}  {0.03}}                                                             & {\color[HTML]{000000}  {0.05}}                                                             & {\color[HTML]{000000}  {0.26}}                                                            \\ \hline
{\color[HTML]{000000} }                                                                                 & {\color[HTML]{000000} }                                                                                      & {\color[HTML]{000000} Default}                            & \cellcolor[HTML]{D28986}{\color[HTML]{000000}  {0.73}}                                        & \cellcolor[HTML]{FFFFFF}{\color[HTML]{000000}  {0.38}}                                             & {\color[HTML]{000000}  {0.66}}                                                                   & \cellcolor[HTML]{D28986}{\color[HTML]{000000}  {0.64}}                                          & {\color[HTML]{000000}  {0.61}}                                                                  & {\color[HTML]{000000}  {0.05}}                                                             & \cellcolor[HTML]{FFFFFF}{\color[HTML]{000000}  {0.14}}                                     & \cellcolor[HTML]{FFFFFF}{\color[HTML]{000000}  {0.18}}                                     & \cellcolor[HTML]{FFFFFF}{\color[HTML]{000000}  {0.28}}                                    \\ \cline{3-12} 
{\color[HTML]{000000} }                                                                                 & {\color[HTML]{000000} }                                                                                      & {\color[HTML]{000000} OP}            & \cellcolor[HTML]{D28986}{\color[HTML]{000000}  {0.71}}                                        & \cellcolor[HTML]{FFFFFF}{\color[HTML]{000000}  {0.36}}                                             & {\color[HTML]{000000}  {0.64}}                                                                   & \cellcolor[HTML]{D28986}{\color[HTML]{000000}  {0.62}}                                          & {\color[HTML]{000000}  {0.60}}                                                                  & {\color[HTML]{000000}  {0.04}}                                                             & {\color[HTML]{000000}  {0.05}}                                                             & {\color[HTML]{000000}  {0.06}}                                                             & {\color[HTML]{000000}  {0.09}}                                                            \\ \cline{3-12} 
{\color[HTML]{000000} }                                                                                 & {\color[HTML]{000000} }                                                                                      & {\color[HTML]{000000} Fairway}                            & \cellcolor[HTML]{FFFFFF}{\color[HTML]{000000}  {0.56}}                                        & {\color[HTML]{000000}  {0.22}}                                                                     & \cellcolor[HTML]{FFCCC9}{\color[HTML]{000000}  {0.57}}                                           & \cellcolor[HTML]{FFCCC9}{\color[HTML]{000000}  {0.58}}                                          & {\color[HTML]{000000}  {0.58}}                                                                  & {\color[HTML]{000000}  {0.03}}                                                             & {\color[HTML]{000000}  {0.03}}                                                             & {\color[HTML]{000000}  {0.06}}                                                             & {\color[HTML]{000000}  {0.08}}                                                            \\ \cline{3-12} 
\multirow{-4}{*}{{\color[HTML]{000000} Compas}}                                                         & \multirow{-4}{*}{{\color[HTML]{000000} Sex}}                                                                 & {\color[HTML]{000000} Fair SMOTE}                         & \cellcolor[HTML]{FFCCC9}{\color[HTML]{000000}  {0.62}}                                        & \cellcolor[HTML]{FFCCC9}{\color[HTML]{000000}  {0.32}}                                             & \cellcolor[HTML]{FFCCC9}{\color[HTML]{000000}  {0.56}}                                           & \cellcolor[HTML]{FFCCC9}{\color[HTML]{000000}  {0.55}}                                          & \cellcolor[HTML]{D28986}{\color[HTML]{000000}  {0.65}}                                          & {\color[HTML]{000000}  {0.02}}                                                             & {\color[HTML]{000000}  {0.05}}                                                             & {\color[HTML]{000000}  {0.08}}                                                             & {\color[HTML]{000000}  {0.09}}                                                            \\ \hline
{\color[HTML]{000000} }                                                                                 & {\color[HTML]{000000} }                                                                                      & {\color[HTML]{000000} Default}                            & \cellcolor[HTML]{D28986}{\color[HTML]{000000}  {0.69}}                                        & \cellcolor[HTML]{FFFFFF}{\color[HTML]{000000}  {0.39}}                                             & {\color[HTML]{000000}  {0.65}}                                                                   & \cellcolor[HTML]{D28986}{\color[HTML]{000000}  {0.64}}                                          & \cellcolor[HTML]{D28986}{\color[HTML]{000000}  {0.68}}                                          & {\color[HTML]{000000}  {0.05}}                                                             & \cellcolor[HTML]{FFFFFF}{\color[HTML]{000000}  {0.11}}                                     & \cellcolor[HTML]{FFFFFF}{\color[HTML]{000000}  {0.12}}                                     & \cellcolor[HTML]{FFFFFF}{\color[HTML]{000000}  {0.21}}                                    \\ \cline{3-12} 
{\color[HTML]{000000} }                                                                                 & {\color[HTML]{000000} }                                                                                      & {\color[HTML]{000000} OP}            & \cellcolor[HTML]{D28986}{\color[HTML]{000000}  {0.68}}                                        & \cellcolor[HTML]{FFCCC9}{\color[HTML]{000000}  {0.33}}                                             & {\color[HTML]{000000}  {0.63}}                                                                   & \cellcolor[HTML]{D28986}{\color[HTML]{000000}  {0.62}}                                          & \cellcolor[HTML]{D28986}{\color[HTML]{000000}  {0.67}}                                          & {\color[HTML]{000000}  {0.03}}                                                             & {\color[HTML]{000000}  {0.06}}                                                             & {\color[HTML]{000000}  {0.06}}                                                             & \cellcolor[HTML]{FFCCC9}{\color[HTML]{000000}  {0.12}}                                    \\ \cline{3-12} 
{\color[HTML]{000000} }                                                                                 & {\color[HTML]{000000} }                                                                                      & {\color[HTML]{000000} Fairway}                            & \cellcolor[HTML]{FFFFFF}{\color[HTML]{000000}  {0.55}}                                        & {\color[HTML]{000000}  {0.21}}                                                                     & \cellcolor[HTML]{FFCCC9}{\color[HTML]{000000}  {0.58}}                                           & \cellcolor[HTML]{FFCCC9}{\color[HTML]{000000}  {0.58}}                                          & {\color[HTML]{000000}  {0.56}}                                                                  & {\color[HTML]{000000}  {0.02}}                                                             & {\color[HTML]{000000}  {0.04}}                                                             & {\color[HTML]{000000}  {0.07}}                                                             & {\color[HTML]{000000}  {0.07}}                                                            \\ \cline{3-12} 
\multirow{-4}{*}{{\color[HTML]{000000} Compas}}                                                         & \multirow{-4}{*}{{\color[HTML]{000000} Race}}                                                                & {\color[HTML]{000000} Fair SMOTE}                         & \cellcolor[HTML]{FFCCC9}{\color[HTML]{000000}  {0.62}}                                        & \cellcolor[HTML]{FFCCC9}{\color[HTML]{000000}  {0.30}}                                             & \cellcolor[HTML]{FFCCC9}{\color[HTML]{000000}  {0.56}}                                           & \cellcolor[HTML]{FFCCC9}{\color[HTML]{000000}  {0.55}}                                          & \cellcolor[HTML]{D28986}{\color[HTML]{000000}  {0.66}}                                          & {\color[HTML]{000000}  {0.01}}                                                             & {\color[HTML]{000000}  {0.05}}                                                             & {\color[HTML]{000000}  {0.06}}                                                             & \cellcolor[HTML]{FFCCC9}{\color[HTML]{000000}  {0.11}}                                    \\ \hline
{\color[HTML]{000000} }                                                                                 & {\color[HTML]{000000} }                                                                                      & {\color[HTML]{000000} Default}                            & \cellcolor[HTML]{D28986}{\color[HTML]{000000}  {0.94}}                                        & \cellcolor[HTML]{FFFFFF}{\color[HTML]{000000}  {0.81}}                                             & {\color[HTML]{000000}  {0.72}}                                                                   & \cellcolor[HTML]{D28986}{\color[HTML]{000000}  {0.76}}                                          & \cellcolor[HTML]{D28986}{\color[HTML]{000000}  {0.82}}                                          & \cellcolor[HTML]{FFFFFF}{\color[HTML]{000000}  {0.11}}                                     & \cellcolor[HTML]{FFFFFF}{\color[HTML]{000000}  {0.08}}                                     & \cellcolor[HTML]{FFFFFF}{\color[HTML]{000000}  {0.14}}                                     & \cellcolor[HTML]{FFFFFF}{\color[HTML]{000000}  {0.15}}                                    \\ \cline{3-12} 
{\color[HTML]{000000} }                                                                                 & {\color[HTML]{000000} }                                                                                      & {\color[HTML]{000000} OP}            & \cellcolor[HTML]{FFCCC9}{\color[HTML]{000000}  {0.75}}                                        & \cellcolor[HTML]{FFCCC9}{\color[HTML]{000000}  {0.73}}                                             & {\color[HTML]{000000}  {0.71}}                                                                   & \cellcolor[HTML]{D28986}{\color[HTML]{000000}  {0.73}}                                          & {\color[HTML]{000000}  {0.71}}                                                                  & {\color[HTML]{000000}  {0.04}}                                                             & {\color[HTML]{000000}  {0.05}}                                                             & {\color[HTML]{000000}  {0.06}}                                                             & {\color[HTML]{000000}  {0.12}}                                                            \\ \cline{3-12} 
{\color[HTML]{000000} }                                                                                 & {\color[HTML]{000000} }                                                                                      & {\color[HTML]{000000} Fairway}                            & \cellcolor[HTML]{FFCCC9}{\color[HTML]{000000}  {0.78}}                                        & \cellcolor[HTML]{FFCCC9}{\color[HTML]{000000}  {0.76}}                                             & {\color[HTML]{000000}  {0.68}}                                                                   & \cellcolor[HTML]{FFCCC9}{\color[HTML]{000000}  {0.65}}                                          & {\color[HTML]{000000}  {0.73}}                                                                  & {\color[HTML]{000000}  {0.05}}                                                             & {\color[HTML]{000000}  {0.04}}                                                             & {\color[HTML]{000000}  {0.07}}                                                             & {\color[HTML]{000000}  {0.11}}                                                            \\ \cline{3-12} 
\multirow{-4}{*}{{\color[HTML]{000000} German Credit}}                                                  & \multirow{-4}{*}{{\color[HTML]{000000} Sex}}                                                                 & {\color[HTML]{000000} Fair SMOTE}                         & \cellcolor[HTML]{FFFFFF}{\color[HTML]{000000}  {0.62}}                                        & {\color[HTML]{000000}  {0.36}}                                                                     & {\color[HTML]{000000}  {0.71}}                                                                   & \cellcolor[HTML]{FFCCC9}{\color[HTML]{000000}  {0.64}}                                          & {\color[HTML]{000000}  {0.71}}                                                                  & {\color[HTML]{000000}  {0.05}}                                                             & {\color[HTML]{000000}  {0.05}}                                                             & {\color[HTML]{000000}  {0.05}}                                                             & {\color[HTML]{000000}  {0.13}}                                                            \\ \hline
{\color[HTML]{000000} }                                                                                 & {\color[HTML]{000000} }                                                                                      & {\color[HTML]{000000} Default}                            & \cellcolor[HTML]{FFCCC9}{\color[HTML]{000000}  {0.25}}                                        & {\color[HTML]{000000}  {0.07}}                                                                     & {\color[HTML]{000000}  {0.7}}                                                                    & \cellcolor[HTML]{D28986}{\color[HTML]{000000}  {0.78}}                                          & {\color[HTML]{000000}  {0.34}}                                                                  & \cellcolor[HTML]{FFFFFF}{\color[HTML]{000000}  {0.05}}                                     & \cellcolor[HTML]{FFFFFF}{\color[HTML]{000000}  {0.08}}                                     & \cellcolor[HTML]{FFFFFF}{\color[HTML]{000000}  {0.06}}                                     & \cellcolor[HTML]{FFFFFF}{\color[HTML]{000000}  {0.45}}                                    \\ \cline{3-12} 
{\color[HTML]{000000} }                                                                                 & {\color[HTML]{000000} }                                                                                      & {\color[HTML]{000000} OP}            & \cellcolor[HTML]{FFCCC9}{\color[HTML]{000000}  {0.28}}                                        & {\color[HTML]{000000}  {0.06}}                                                                     & {\color[HTML]{000000}  {0.65}}                                                                   & \cellcolor[HTML]{FFCCC9}{\color[HTML]{000000}  {0.70}}                                          & {\color[HTML]{000000}  {0.32}}                                                                  & {\color[HTML]{000000}  {0.01}}                                                             & {\color[HTML]{000000}  {0.02}}                                                             & {\color[HTML]{000000}  {0.03}}                                                             & \cellcolor[HTML]{FFCCC9}{\color[HTML]{000000}  {0.09}}                                    \\ \cline{3-12} 
{\color[HTML]{000000} }                                                                                 & {\color[HTML]{000000} }                                                                                      & {\color[HTML]{000000} Fairway}                            & \cellcolor[HTML]{FFFFFF}{\color[HTML]{000000}  {0.21}}                                        & {\color[HTML]{000000}  {0.04}}                                                                     & {\color[HTML]{000000}  {0.67}}                                                                   & \cellcolor[HTML]{FFCCC9}{\color[HTML]{000000}  {0.67}}                                          & {\color[HTML]{000000}  {0.33}}                                                                  & {\color[HTML]{000000}  {0.01}}                                                             & {\color[HTML]{000000}  {0.04}}                                                             & {\color[HTML]{000000}  {0.03}}                                                             & \cellcolor[HTML]{FFCCC9}{\color[HTML]{000000}  {0.12}}                                    \\ \cline{3-12} 
\multirow{-4}{*}{{\color[HTML]{000000} Default Credit}}                                                 & \multirow{-4}{*}{{\color[HTML]{000000} Sex}}                                                                 & {\color[HTML]{000000} Fair SMOTE}                         & \cellcolor[HTML]{D28986}{\color[HTML]{000000}  {0.58}}                                        & \cellcolor[HTML]{FFCCC9}{\color[HTML]{000000}  {0.26}}                                             & \cellcolor[HTML]{FFCCC9}{\color[HTML]{000000}  {0.39}}                                           & \cellcolor[HTML]{FFCCC9}{\color[HTML]{000000}  {0.68}}                                          & \cellcolor[HTML]{D28986}{\color[HTML]{000000}  {0.44}}                                          & {\color[HTML]{000000}  {0.02}}                                                             & {\color[HTML]{000000}  {0.03}}                                                             & {\color[HTML]{000000}  {0.05}}                                                             &  {0.03}                                                                                   \\ \hline
{\color[HTML]{000000} }                                                                                 & {\color[HTML]{000000} }                                                                                      & {\color[HTML]{000000} Default}                            & \cellcolor[HTML]{D28986}{\color[HTML]{000000}  {0.72}}                                        & {\color[HTML]{000000}  {0.2}}                                                                      & {\color[HTML]{000000}  {0.74}}                                                                   & \cellcolor[HTML]{D28986}{\color[HTML]{000000}  {0.72}}                                          & \cellcolor[HTML]{D28986}{\color[HTML]{000000}  {0.74}}                                          & \cellcolor[HTML]{FFFFFF}{\color[HTML]{000000}  {0.11}}                                     & \cellcolor[HTML]{FFFFFF}{\color[HTML]{000000}  {0.13}}                                     & \cellcolor[HTML]{FFFFFF}{\color[HTML]{000000}  {0.32}}                                     & \cellcolor[HTML]{FFFFFF}{\color[HTML]{000000}  {0.44}}                                    \\ \cline{3-12} 
{\color[HTML]{000000} }                                                                                 & {\color[HTML]{000000} }                                                                                      & {\color[HTML]{000000} OP}            & \cellcolor[HTML]{FFCCC9}{\color[HTML]{000000}  {0.69}}                                        & {\color[HTML]{000000}  {0.19}}                                                                     & {\color[HTML]{000000}  {0.72}}                                                                   & \cellcolor[HTML]{FFCCC9}{\color[HTML]{000000}  {0.69}}                                          & {\color[HTML]{000000}  {0.69}}                                                                  & {\color[HTML]{000000}  {0.04}}                                                             & {\color[HTML]{000000}  {0.06}}                                                             & {\color[HTML]{000000}  {0.09}}                                                             & {\color[HTML]{000000}  {0.14}}                                                            \\ \cline{3-12} 
{\color[HTML]{000000} }                                                                                 & {\color[HTML]{000000} }                                                                                      & {\color[HTML]{000000} Fairway}                            & \cellcolor[HTML]{FFCCC9}{\color[HTML]{000000}  {0.65}}                                        & {\color[HTML]{000000}  {0.21}}                                                                     & \cellcolor[HTML]{FFCCC9}{\color[HTML]{000000}  {0.69}}                                           & \cellcolor[HTML]{FFCCC9}{\color[HTML]{000000}  {0.67}}                                          & {\color[HTML]{000000}  {0.68}}                                                                  & {\color[HTML]{000000}  {0.05}}                                                             & {\color[HTML]{000000}  {0.05}}                                                             & {\color[HTML]{000000}  {0.04}}                                                             & {\color[HTML]{000000}  {0.18}}                                                            \\ \cline{3-12} 
\multirow{-4}{*}{{\color[HTML]{000000} Heart Health}}                                                   & \multirow{-4}{*}{{\color[HTML]{000000} Age}}                                                                 & {\color[HTML]{000000} Fair SMOTE}                         & \cellcolor[HTML]{FFCCC9}{\color[HTML]{000000}  {0.66}}                                        & {\color[HTML]{000000}  {0.20}}                                                                     & \cellcolor[HTML]{FFCCC9}{\color[HTML]{000000}  {0.69}}                                           & \cellcolor[HTML]{FFCCC9}{\color[HTML]{000000}  {0.68}}                                          & {\color[HTML]{000000}  {0.66}}                                                                  & {\color[HTML]{000000}  {0.08}}                                                             & {\color[HTML]{000000}  {0.07}}                                                             & {\color[HTML]{000000}  {0.08}}                                                             & {\color[HTML]{000000}  {0.20}}                                                            \\ \hline
{\color[HTML]{000000} }                                                                                 & {\color[HTML]{000000} }                                                                                      & {\color[HTML]{000000} Default}                            & \cellcolor[HTML]{FFCCC9}{\color[HTML]{000000}  {0.73}}                                        & \cellcolor[HTML]{FFCCC9}{\color[HTML]{000000}  {0.21}}                                             & {\color[HTML]{000000}  {0.76}}                                                                   & \cellcolor[HTML]{D28986}{\color[HTML]{000000}  {0.77}}                                          & \cellcolor[HTML]{D28986}{\color[HTML]{000000}  {0.77}}                                          & \cellcolor[HTML]{FFFFFF}{\color[HTML]{000000}  {0.14}}                                     & \cellcolor[HTML]{FFFFFF}{\color[HTML]{000000}  {0.22}}                                     & \cellcolor[HTML]{FFFFFF}{\color[HTML]{000000}  {0.24}}                                     & \cellcolor[HTML]{FFFFFF}{\color[HTML]{000000}  {0.21}}                                    \\ \cline{3-12} 
{\color[HTML]{000000} }                                                                                 & {\color[HTML]{000000} }                                                                                      & {\color[HTML]{000000} OP}            & \cellcolor[HTML]{FFCCC9}{\color[HTML]{000000}  {0.72}}                                        & \cellcolor[HTML]{FFCCC9}{\color[HTML]{000000}  {0.20}}                                             & \cellcolor[HTML]{FFCCC9}{\color[HTML]{000000}  {0.74}}                                           & \cellcolor[HTML]{D28986}{\color[HTML]{000000}  {0.75}}                                          & {\color[HTML]{000000}  {0.75}}                                                                  & {\color[HTML]{000000}  {0.05}}                                                             & {\color[HTML]{000000}  {0.05}}                                                             & {\color[HTML]{000000}  {0.02}}                                                             & {\color[HTML]{000000}  {0.04}}                                                            \\ \cline{3-12} 
{\color[HTML]{000000} }                                                                                 & {\color[HTML]{000000} }                                                                                      & {\color[HTML]{000000} Fairway}                            & \cellcolor[HTML]{FFCCC9}{\color[HTML]{000000}  {0.71}}                                        & {\color[HTML]{000000}  {0.17}}                                                                     & \cellcolor[HTML]{FFCCC9}{\color[HTML]{000000}  {0.73}}                                           & \cellcolor[HTML]{FFCCC9}{\color[HTML]{000000}  {0.71}}                                          & {\color[HTML]{000000}  {0.73}}                                                                  & {\color[HTML]{000000}  {0.04}}                                                             & {\color[HTML]{000000}  {0.03}}                                                             & {\color[HTML]{000000}  {0.03}}                                                             & {\color[HTML]{000000}  {0.05}}                                                            \\ \cline{3-12} 
\multirow{-4}{*}{{\color[HTML]{000000} Bank Marketing}}                                                 & \multirow{-4}{*}{{\color[HTML]{000000} Age}}                                                                 & {\color[HTML]{000000} Fair SMOTE}                         & \cellcolor[HTML]{D28986}{\color[HTML]{000000}  {0.76}}                                        & {\color[HTML]{000000}  {0.18}}                                                                     & \cellcolor[HTML]{FFCCC9}{\color[HTML]{000000}  {0.72}}                                           & \cellcolor[HTML]{FFCCC9}{\color[HTML]{000000}  {0.72}}                                          & {\color[HTML]{000000}  {0.74}}                                                                  & {\color[HTML]{000000}  {0.05}}                                                             & {\color[HTML]{000000}  {0.07}}                                                             & {\color[HTML]{000000}  {0.05}}                                                             & {\color[HTML]{000000}  {0.03}}                                                            \\ \hline
{\color[HTML]{000000} }                                                                                 & {\color[HTML]{000000} }                                                                                      & {\color[HTML]{000000} Default}                            & \cellcolor[HTML]{D28986}{\color[HTML]{000000}  {0.31}}                                        & {\color[HTML]{000000}  {0.18}}                                                                     & {\color[HTML]{000000}  {0.28}}                                                                   & \cellcolor[HTML]{D28986}{\color[HTML]{000000}  {0.86}}                                          & \cellcolor[HTML]{D28986}{\color[HTML]{000000}  {0.29}}                                          & \cellcolor[HTML]{FFFFFF}{\color[HTML]{000000}  {0.07}}                                     & \cellcolor[HTML]{FFFFFF}{\color[HTML]{000000}  {0.06}}                                     & \cellcolor[HTML]{FFFFFF}{\color[HTML]{000000}  {0.08}}                                     & \cellcolor[HTML]{FFFFFF}{\color[HTML]{000000}  {0.59}}                                    \\ \cline{3-12} 
{\color[HTML]{000000} }                                                                                 & {\color[HTML]{000000} }                                                                                      & {\color[HTML]{000000} OP}            & \cellcolor[HTML]{D28986}{\color[HTML]{000000}  {0.3}}                                         & {\color[HTML]{000000}  {0.17}}                                                                     & {\color[HTML]{000000}  {0.30}}                                                                   & \cellcolor[HTML]{D28986}{\color[HTML]{000000}  {0.83}}                                          & \cellcolor[HTML]{D28986}{\color[HTML]{000000}  {0.31}}                                          & {\color[HTML]{000000}  {0.02}}                                                             & {\color[HTML]{000000}  {0.04}}                                                             & {\color[HTML]{000000}  {0.04}}                                                             & {\color[HTML]{000000}  {0.16}}                                                            \\ \cline{3-12} 
{\color[HTML]{000000} }                                                                                 & {\color[HTML]{000000} }                                                                                      & {\color[HTML]{000000} Fairway}                            & \cellcolor[HTML]{FFCCC9}{\color[HTML]{000000}  {0.28}}                                        & \cellcolor[HTML]{FFCCC9}{\color[HTML]{000000}  {0.12}}                                             & \cellcolor[HTML]{FFCCC9}{\color[HTML]{000000}  {0.25}}                                           & \cellcolor[HTML]{FFFFFF}{\color[HTML]{000000}  {0.65}}                                          &  {0.26}                                                                                         & {\color[HTML]{000000}  {0.02}}                                                             & {\color[HTML]{000000}  {0.04}}                                                             & {\color[HTML]{000000}  {0.03}}                                                             & {\color[HTML]{000000}  {013}}                                                             \\ \cline{3-12} 
\multirow{-4}{*}{{\color[HTML]{000000} Home Credit}}                                                    & \multirow{-4}{*}{{\color[HTML]{000000} Sex}}                                                                 & {\color[HTML]{000000} Fair SMOTE}                         & \cellcolor[HTML]{D28986}{\color[HTML]{000000}  {0.33}}                                        & {\color[HTML]{000000}  {0.18}}                                                                     & {\color[HTML]{000000}  {0.31}}                                                                   & \cellcolor[HTML]{FFCCC9}{\color[HTML]{000000}  {0.75}}                                          & \cellcolor[HTML]{D28986}{\color[HTML]{000000}  {0.32}}                                          & {\color[HTML]{000000}  {0.03}}                                                             & {\color[HTML]{000000}  {0.02}}                                                             & {\color[HTML]{000000}  {0.03}}                                                             & {\color[HTML]{000000}  {0.15}}                                                            \\ \hline
{\color[HTML]{000000} }                                                                                 & {\color[HTML]{000000} }                                                                                      & {\color[HTML]{000000} Default}                            & \cellcolor[HTML]{FFCCC9}{\color[HTML]{000000}  {0.81}}                                        & {\color[HTML]{000000}  {0.06}}                                                                     &  {0.85}                                                                                          & \cellcolor[HTML]{D28986} {0.88}                                                                 & \cellcolor[HTML]{D28986}{\color[HTML]{000000}  {0.83}}                                          & \cellcolor[HTML]{FFFFFF}{\color[HTML]{000000}  {0.06}}                                     & \cellcolor[HTML]{FFFFFF}{\color[HTML]{000000}  {0.05}}                                     & \cellcolor[HTML]{FFFFFF}{\color[HTML]{000000}  {0.06}}                                     & \cellcolor[HTML]{FFFFFF}{\color[HTML]{000000}  {0.12}}                                    \\ \cline{3-12} 
{\color[HTML]{000000} }                                                                                 & {\color[HTML]{000000} }                                                                                      & {\color[HTML]{000000} OP}            & \cellcolor[HTML]{FFCCC9}{\color[HTML]{000000}  {0.79}}                                        & {\color[HTML]{000000}  {0.06}}                                                                     &  {0.83}                                                                                          & \cellcolor[HTML]{D28986} {0.83}                                                                 & \cellcolor[HTML]{D28986}{\color[HTML]{000000}  {0.82}}                                          & {\color[HTML]{000000}  {0.03}}                                                             & {\color[HTML]{000000}  {0.02}}                                                             & {\color[HTML]{000000}  {0.03}}                                                             & {\color[HTML]{000000}  {0.06}}                                                            \\ \cline{3-12} 
{\color[HTML]{000000} }                                                                                 & {\color[HTML]{000000} }                                                                                      & {\color[HTML]{000000} Fairway}                            & \cellcolor[HTML]{FFFFFF}{\color[HTML]{000000}  {0.76}}                                        & {\color[HTML]{000000}  {0.05}}                                                                     &  {0.81}                                                                                          & \cellcolor[HTML]{D28986} {0.84}                                                                 & \cellcolor[HTML]{D28986}{\color[HTML]{000000}  {0.84}}                                          & {\color[HTML]{000000}  {0.03}}                                                             & {\color[HTML]{000000}  {0.02}}                                                             & {\color[HTML]{000000}  {0.04}}                                                             & {\color[HTML]{000000}  {0.07}}                                                            \\ \cline{3-12} 
\multirow{-4}{*}{{\color[HTML]{000000} \begin{tabular}[c]{@{}c@{}}Student\\ Performance\end{tabular}}}  & \multirow{-4}{*}{{\color[HTML]{000000} Sex}}                                                                 & {\color[HTML]{000000} Fair SMOTE}                         & \cellcolor[HTML]{D28986}{\color[HTML]{000000}  {0.91}}                                        & \cellcolor[HTML]{FFCCC9}{\color[HTML]{000000}  {0.10}}                                             &  {0.84}                                                                                          & \cellcolor[HTML]{D28986} {0.87}                                                                 & \cellcolor[HTML]{D28986}{\color[HTML]{000000}  {0.86}}                                          & {\color[HTML]{000000}  {0.04}}                                                             & {\color[HTML]{000000}  {0.04}}                                                             & {\color[HTML]{000000}  {0.04}}                                                             & {\color[HTML]{000000}  {0.08}}                                                            \\ \hline
{\color[HTML]{000000} }                                                                                 & {\color[HTML]{000000} }                                                                                      & {\color[HTML]{000000} Default}                            & \cellcolor[HTML]{FFCCC9}{\color[HTML]{000000}  {0.36}}                                        & {\color[HTML]{000000}  {0.03}}                                                                     &  {0.68}                                                                                          & \cellcolor[HTML]{D28986} {0.85}                                                                 & {\color[HTML]{000000}  {0.45}}                                                                  & {\color[HTML]{000000}  {0.04}}                                                             & \cellcolor[HTML]{FFFFFF}{\color[HTML]{000000}  {0.05}}                                     & \cellcolor[HTML]{FFFFFF}{\color[HTML]{000000}  {0.08}}                                     & \cellcolor[HTML]{FFFFFF}{\color[HTML]{000000}  {0.36}}                                    \\ \cline{3-12} 
{\color[HTML]{000000} }                                                                                 & {\color[HTML]{000000} }                                                                                      & {\color[HTML]{000000} OP}            & \cellcolor[HTML]{FFCCC9}{\color[HTML]{000000}  {0.35}}                                        & {\color[HTML]{000000}  {0.04}}                                                                     &  {0.66}                                                                                          & \cellcolor[HTML]{D28986} {0.83}                                                                 & {\color[HTML]{000000}  {0.44}}                                                                  & {\color[HTML]{000000}  {0.04}}                                                             & {\color[HTML]{000000}  {0.02}}                                                             & {\color[HTML]{000000}  {0.06}}                                                             & {\color[HTML]{000000}  {0.15}}                                                            \\ \cline{3-12} 
{\color[HTML]{000000} }                                                                                 & {\color[HTML]{000000} }                                                                                      & {\color[HTML]{000000} Fairway}                            & \cellcolor[HTML]{FFCCC9}{\color[HTML]{000000}  {0.35}}                                        & {\color[HTML]{000000}  {0.04}}                                                                     & \cellcolor[HTML]{FFCCC9}{\color[HTML]{000000}  {0.42}}                                           & \cellcolor[HTML]{FFCCC9}{\color[HTML]{000000}  {0.77}}                                          & \cellcolor[HTML]{FFFFFF}{\color[HTML]{000000}  {0.41}}                                          & {\color[HTML]{000000}  {0.03}}                                                             & {\color[HTML]{000000}  {0.02}}                                                             & {\color[HTML]{000000}  {0.04}}                                                             & {\color[HTML]{000000}  {0.12}}                                                            \\ \cline{3-12} 
\multirow{-4}{*}{{\color[HTML]{000000} MEPS - 15}}                                                      & \multirow{-4}{*}{{\color[HTML]{000000} Race}}                                                                & {\color[HTML]{000000} Fair SMOTE}                         & \cellcolor[HTML]{D28986}{\color[HTML]{000000}  {0.68}}                                        & \cellcolor[HTML]{FFCCC9}{\color[HTML]{000000}  {0.22}}                                             & \cellcolor[HTML]{FFCCC9}{\color[HTML]{000000}  {0.41}}                                           & \cellcolor[HTML]{FFCCC9}{\color[HTML]{000000}  {0.77}}                                          & \cellcolor[HTML]{D28986}{\color[HTML]{000000}  {0.53}}                                          & {\color[HTML]{000000}  {0.02}}                                                             & {\color[HTML]{000000}  {0.02}}                                                             & {\color[HTML]{000000}  {0.05}}                                                             & {\color[HTML]{000000}  {0.15}}                                                            \\ \hline
{\color[HTML]{000000} }                                                                                 & {\color[HTML]{000000} }                                                                                      & {\color[HTML]{000000} Default}                            & \cellcolor[HTML]{FFCCC9}{\color[HTML]{000000}  {0.35}}                                        & {\color[HTML]{000000}  {0.05}}                                                                     & {\color[HTML]{000000}  {0.65}}                                                                   & \cellcolor[HTML]{D28986}{\color[HTML]{000000}  {0.85}}                                          & {\color[HTML]{000000}  {0.44}}                                                                  & {\color[HTML]{000000}  {0.04}}                                                             & \cellcolor[HTML]{FFFFFF}{\color[HTML]{000000}  {0.1}}                                      & \cellcolor[HTML]{FFFFFF}{\color[HTML]{000000}  {0.07}}                                     & \cellcolor[HTML]{FFFFFF}{\color[HTML]{000000}  {0.43}}                                    \\ \cline{3-12} 
{\color[HTML]{000000} }                                                                                 & {\color[HTML]{000000} }                                                                                      & {\color[HTML]{000000} OP}            & \cellcolor[HTML]{FFCCC9}{\color[HTML]{000000}  {0.34}}                                        & {\color[HTML]{000000}  {0.04}}                                                                     & {\color[HTML]{000000}  {0.65}}                                                                   & \cellcolor[HTML]{D28986}{\color[HTML]{000000}  {0.83}}                                          & {\color[HTML]{000000}  {0.44}}                                                                  & {\color[HTML]{000000}  {0.05}}                                                             & {\color[HTML]{000000}  {0.02}}                                                             & {\color[HTML]{000000}  {0.05}}                                                             & {\color[HTML]{000000}  {0.12}}                                                            \\ \cline{3-12} 
{\color[HTML]{000000} }                                                                                 & {\color[HTML]{000000} }                                                                                      & {\color[HTML]{000000} Fairway}                            & \cellcolor[HTML]{FFCCC9}{\color[HTML]{000000}  {0.32}}                                        & {\color[HTML]{000000}  {0.04}}                                                                     & \cellcolor[HTML]{FFCCC9}{\color[HTML]{000000}  {0.55}}                                           & \cellcolor[HTML]{FFCCC9}{\color[HTML]{000000}  {0.76}}                                          & {\color[HTML]{000000}  {0.42}}                                                                  & {\color[HTML]{000000}  {0.03}}                                                             & {\color[HTML]{000000}  {0.02}}                                                             & {\color[HTML]{000000}  {0.04}}                                                             & {\color[HTML]{000000}  {0.16}}                                                            \\ \cline{3-12} 
\multirow{-4}{*}{{\color[HTML]{000000} MEPS - 16}}                                                      & \multirow{-4}{*}{{\color[HTML]{000000} Sex}}                                                                 & {\color[HTML]{000000} Fair SMOTE}                         & \cellcolor[HTML]{D28986}{\color[HTML]{000000}  {0.66}}                                        & \cellcolor[HTML]{FFCCC9}{\color[HTML]{000000}  {0.2}}                                              & \cellcolor[HTML]{FFFFFF}{\color[HTML]{000000}  {0.41}}                                           & \cellcolor[HTML]{FFCCC9}{\color[HTML]{000000}  {0.77}}                                          & \cellcolor[HTML]{D28986}{\color[HTML]{000000}  {0.51}}                                          & {\color[HTML]{000000}  {0.01}}                                                             & {\color[HTML]{000000}  {0.03}}                                                             & {\color[HTML]{000000}  {0.04}}                                                             & {\color[HTML]{000000}  {0.17}}                                                            \\ \hline
\end{tabular}}
\end{table*}

\bi
\item \textbf{Data distribution} - The training data should be almost balanced based on class and protected attribute.
\item \textbf{Data label} - Every data point should go through \textit{situation testing} to see whether label has bias or not.
\ei
If we find bias in the training data, we should apply Fair-SMOTE to remove that and get fair outcomes. Thus, the answer for RQ1 is ``\textbf{Yes, we can find bias by just looking at the training data}''

\begin{table*}[h]
\caption{RQ3, RQ4 results. Summarized information of comparing Fair-SMOTE with SMOTE\cite{Chawla_2002}, Fairway\cite{Chakraborty_2020} \& Optimized Pre-processing\cite{NIPS2017_6988} based on results of 10 datasets and 3 learners (LSR, RF, SVM). Number of wins, ties, and losses are calculated based on Scott-Knott ranks for each metric. Highlighted cells show Fair-SMOTE significantly outperforming others.}
\label{Fair-SMOTE_vs_others}
\small
\begin{tabular}{
>{\columncolor[HTML]{C0C0C0}}c 
>{\columncolor[HTML]{C0C0C0}}c cccccccccc}
\cline{3-12}
\multicolumn{1}{l}{\cellcolor[HTML]{C0C0C0}} &                                                                 & \cellcolor[HTML]{C0C0C0}\textbf{Recall}         & \cellcolor[HTML]{C0C0C0}\textbf{False alarm} & \cellcolor[HTML]{C0C0C0}\textbf{Precision} & \cellcolor[HTML]{C0C0C0}\textbf{Accuracy} & \cellcolor[HTML]{C0C0C0}\textbf{F1 Score}       & \cellcolor[HTML]{C0C0C0}\textbf{AOD}            & \cellcolor[HTML]{C0C0C0}\textbf{EOD}            & \cellcolor[HTML]{C0C0C0}\textbf{SPD}            & \cellcolor[HTML]{C0C0C0}\textbf{DI}             & \cellcolor[HTML]{C0C0C0}\textbf{Total} \\ \cline{3-12} 
\multicolumn{1}{l}{\cellcolor[HTML]{C0C0C0}} & \multicolumn{11}{c}{\cellcolor[HTML]{C0C0C0}\textbf{SMOTE vs Fair-SMOTE}}                                                                                                                                                                                                                                                                                                                                                                                                                                                                                    \\ \cline{3-12} 
1                                            & \multicolumn{1}{c|}{\cellcolor[HTML]{C0C0C0}\textbf{Win}}       & \multicolumn{1}{c|}{4}                          & \multicolumn{1}{c|}{4}                       & \multicolumn{1}{c|}{1}                     & \multicolumn{1}{c|}{6}                    & \multicolumn{1}{c|}{3}                          & \multicolumn{1}{c|}{\cellcolor[HTML]{FFABA7}33} & \multicolumn{1}{c|}{\cellcolor[HTML]{FFABA7}33} & \multicolumn{1}{c|}{\cellcolor[HTML]{FFABA7}34} & \multicolumn{1}{c|}{\cellcolor[HTML]{FFABA7}32} & \multicolumn{1}{c|}{150}               \\ \cline{3-12} 
2                                            & \multicolumn{1}{c|}{\cellcolor[HTML]{C0C0C0}\textbf{Tie}}       & \multicolumn{1}{c|}{25}                         & \multicolumn{1}{c|}{27}                      & \multicolumn{1}{c|}{29}                    & \multicolumn{1}{c|}{28}                   & \multicolumn{1}{c|}{30}                         & \multicolumn{1}{c|}{\cellcolor[HTML]{FFABA7}2}  & \multicolumn{1}{c|}{\cellcolor[HTML]{FFABA7}3}  & \multicolumn{1}{c|}{\cellcolor[HTML]{FFABA7}2}  & \multicolumn{1}{c|}{\cellcolor[HTML]{FFABA7}2}  & \multicolumn{1}{c|}{148}               \\ \cline{3-12} 
3                                            & \multicolumn{1}{c|}{\cellcolor[HTML]{C0C0C0}\textbf{Loss}}      & \multicolumn{1}{c|}{7}                          & \multicolumn{1}{c|}{5}                       & \multicolumn{1}{c|}{6}                     & \multicolumn{1}{c|}{2}                    & \multicolumn{1}{c|}{3}                          & \multicolumn{1}{c|}{1}                          & \multicolumn{1}{c|}{0}                          & \multicolumn{1}{c|}{0}                          & \multicolumn{1}{c|}{2}                          & \multicolumn{1}{c|}{26}                \\ \cline{3-12} 
4                                            & \multicolumn{1}{c|}{\cellcolor[HTML]{C0C0C0}\textbf{Win + Tie}} & \multicolumn{1}{c|}{29}                         & \multicolumn{1}{c|}{31}                      & \multicolumn{1}{c|}{30}                    & \multicolumn{1}{c|}{34}                   & \multicolumn{1}{c|}{33}                         & \multicolumn{1}{c|}{\cellcolor[HTML]{FFABA7}35} & \multicolumn{1}{c|}{\cellcolor[HTML]{FFABA7}36} & \multicolumn{1}{c|}{\cellcolor[HTML]{FFABA7}36} & \multicolumn{1}{c|}{\cellcolor[HTML]{FFABA7}34} & \multicolumn{1}{c|}{298/324}           \\ \cline{3-12} 
                                             & \multicolumn{11}{c}{\cellcolor[HTML]{C0C0C0}\textbf{Fairway vs Fair-SMOTE}}                                                                                                                                                                                                                                                                                                                                                                                                                                                                                  \\ \cline{3-12} 
5                                            & \multicolumn{1}{c|}{\cellcolor[HTML]{C0C0C0}\textbf{Win}}       & \multicolumn{1}{c|}{\cellcolor[HTML]{FFABA7}14} & \multicolumn{1}{c|}{4}                       & \multicolumn{1}{c|}{6}                     & \multicolumn{1}{c|}{5}                    & \multicolumn{1}{c|}{\cellcolor[HTML]{FFABA7}20} & \multicolumn{1}{c|}{3}                          & \multicolumn{1}{c|}{2}                          & \multicolumn{1}{c|}{3}                          & \multicolumn{1}{c|}{4}                          & \multicolumn{1}{c|}{61}                \\ \cline{3-12} 
6                                            & \multicolumn{1}{c|}{\cellcolor[HTML]{C0C0C0}\textbf{Tie}}       & \multicolumn{1}{c|}{\cellcolor[HTML]{FFABA7}20} & \multicolumn{1}{c|}{20}                      & \multicolumn{1}{c|}{27}                    & \multicolumn{1}{c|}{28}                   & \multicolumn{1}{c|}{\cellcolor[HTML]{FFABA7}14} & \multicolumn{1}{c|}{30}                         & \multicolumn{1}{c|}{31}                         & \multicolumn{1}{c|}{32}                         & \multicolumn{1}{c|}{31}                         & \multicolumn{1}{c|}{233}               \\ \cline{3-12} 
7                                            & \multicolumn{1}{c|}{\cellcolor[HTML]{C0C0C0}\textbf{Loss}}      & \multicolumn{1}{c|}{2}                          & \multicolumn{1}{c|}{12}                      & \multicolumn{1}{c|}{3}                     & \multicolumn{1}{c|}{3}                    & \multicolumn{1}{c|}{2}                          & \multicolumn{1}{c|}{3}                          & \multicolumn{1}{c|}{3}                          & \multicolumn{1}{c|}{1}                          & \multicolumn{1}{c|}{1}                          & \multicolumn{1}{c|}{28}                \\ \cline{3-12} 
8                                            & \multicolumn{1}{c|}{\cellcolor[HTML]{C0C0C0}\textbf{Win + Tie}} & \multicolumn{1}{c|}{\cellcolor[HTML]{FFABA7}34} & \multicolumn{1}{c|}{24}                      & \multicolumn{1}{c|}{33}                    & \multicolumn{1}{c|}{33}                   & \multicolumn{1}{c|}{\cellcolor[HTML]{FFABA7}34} & \multicolumn{1}{c|}{33}                         & \multicolumn{1}{c|}{33}                         & \multicolumn{1}{c|}{35}                         & \multicolumn{1}{c|}{35}                         & \multicolumn{1}{c|}{294/324}           \\ \cline{3-12} 
                                             & \multicolumn{11}{c}{\cellcolor[HTML]{C0C0C0}\textbf{Optimized Pre-processing vs Fair-SMOTE}}                                                                                                                                                                                                                                                                                                                                                                                                                                                                 \\ \cline{3-12} 
9                                            & \multicolumn{1}{c|}{\cellcolor[HTML]{C0C0C0}\textbf{Win}}       & \multicolumn{1}{c|}{\cellcolor[HTML]{FFABA7}10} & \multicolumn{1}{c|}{7}                       & \multicolumn{1}{c|}{4}                     & \multicolumn{1}{c|}{3}                    & \multicolumn{1}{c|}{\cellcolor[HTML]{FFABA7}12} & \multicolumn{1}{c|}{1}                          & \multicolumn{1}{c|}{2}                          & \multicolumn{1}{c|}{2}                          & \multicolumn{1}{c|}{3}                          & \multicolumn{1}{c|}{44}                \\ \cline{3-12} 
10                                           & \multicolumn{1}{c|}{\cellcolor[HTML]{C0C0C0}\textbf{Tie}}       & \multicolumn{1}{c|}{\cellcolor[HTML]{FFABA7}21} & \multicolumn{1}{c|}{22}                      & \multicolumn{1}{c|}{26}                    & \multicolumn{1}{c|}{30}                   & \multicolumn{1}{c|}{\cellcolor[HTML]{FFABA7}20} & \multicolumn{1}{c|}{34}                         & \multicolumn{1}{c|}{33}                         & \multicolumn{1}{c|}{32}                         & \multicolumn{1}{c|}{31}                         & \multicolumn{1}{c|}{249}               \\ \cline{3-12} 
11                                           & \multicolumn{1}{c|}{\cellcolor[HTML]{C0C0C0}\textbf{Loss}}      & \multicolumn{1}{c|}{5}                          & \multicolumn{1}{c|}{7}                       & \multicolumn{1}{c|}{6}                     & \multicolumn{1}{c|}{3}                    & \multicolumn{1}{c|}{4}                          & \multicolumn{1}{c|}{1}                          & \multicolumn{1}{c|}{1}                          & \multicolumn{1}{c|}{2}                          & \multicolumn{1}{c|}{2}                          & \multicolumn{1}{c|}{31}                \\ \cline{3-12} 
12                                           & \multicolumn{1}{c|}{\cellcolor[HTML]{C0C0C0}\textbf{Win + Tie}} & \multicolumn{1}{c|}{\cellcolor[HTML]{FFABA7}31} & \multicolumn{1}{c|}{29}                      & \multicolumn{1}{c|}{30}                    & \multicolumn{1}{c|}{33}                   & \multicolumn{1}{c|}{\cellcolor[HTML]{FFABA7}32} & \multicolumn{1}{c|}{35}                         & \multicolumn{1}{c|}{35}                         & \multicolumn{1}{c|}{34}                         & \multicolumn{1}{c|}{34}                         & \multicolumn{1}{c|}{293/324}           \\ \cline{3-12} 
\end{tabular}
\end{table*}

\begin{table*}[]
\small
\caption{RQ5 results. Fair-SMOTE reducing bias for `sex' and `race' simultaneously (Adult dataset). Best cells are highlighted.}
\label{two_protected}
\begin{tabular}{|c|c|c|c|c|c|c|c|c|c|c|}
\hline
\rowcolor[HTML]{C0C0C0} 
\textbf{}                                            & \begin{tabular}[c]{@{}c@{}}Protected\\ attribute\end{tabular} & Recall(+)                                                             & False alarm(-)                                                        & Precision(+)                                                          & Accuracy(+)                                                           & F1 Score(+)                                                           & AOD(-)                                              & EOD(-)                                              & SPD(-)                                              & DI(-)                                               \\ \hline
\cellcolor[HTML]{C0C0C0}                             & Sex                                                           & \cellcolor[HTML]{FFFFFF}{\color[HTML]{000000} }                       & \cellcolor[HTML]{FFABA7}{\color[HTML]{000000} }                       & \cellcolor[HTML]{FFABA7}{\color[HTML]{000000} }                       & \cellcolor[HTML]{FFABA7}{\color[HTML]{000000} }                       & \cellcolor[HTML]{FFFFFF}{\color[HTML]{000000} }                       & \cellcolor[HTML]{FFFFFF}{\color[HTML]{000000} 0.12} & \cellcolor[HTML]{FFFFFF}{\color[HTML]{000000} 0.24} & \cellcolor[HTML]{FFFFFF}{\color[HTML]{000000} 0.21} & \cellcolor[HTML]{FFFFFF}{\color[HTML]{000000} 0.56} \\ \cline{2-2} \cline{8-11} 
\multirow{-2}{*}{\cellcolor[HTML]{C0C0C0}Default}    & Race                                                          & \multirow{-2}{*}{\cellcolor[HTML]{FFFFFF}{\color[HTML]{000000} 0.42}} & \multirow{-2}{*}{\cellcolor[HTML]{FFABA7}{\color[HTML]{000000} 0.07}} & \multirow{-2}{*}{\cellcolor[HTML]{FFABA7}{\color[HTML]{000000} 0.69}} & \multirow{-2}{*}{\cellcolor[HTML]{FFABA7}{\color[HTML]{000000} 0.83}} & \multirow{-2}{*}{\cellcolor[HTML]{FFFFFF}{\color[HTML]{000000} 0.52}} & \cellcolor[HTML]{FFFFFF}{\color[HTML]{000000} 0.06} & \cellcolor[HTML]{FFFFFF}{\color[HTML]{000000} 0.15} & \cellcolor[HTML]{FFFFFF}{\color[HTML]{000000} 0.16} & \cellcolor[HTML]{FFFFFF}{\color[HTML]{000000} 0.52} \\ \hline
\cellcolor[HTML]{C0C0C0}                             & Sex                                                           & \cellcolor[HTML]{FFABA7}{\color[HTML]{000000} }                       & \cellcolor[HTML]{FFFFFF}{\color[HTML]{000000} }                       & \cellcolor[HTML]{FFFFFF}{\color[HTML]{000000} }                       & \cellcolor[HTML]{FFFFFF}{\color[HTML]{000000} }                       & \cellcolor[HTML]{FFABA7}{\color[HTML]{000000} }                       & \cellcolor[HTML]{FFABA7}{\color[HTML]{000000} 0.02} & \cellcolor[HTML]{FFABA7}{\color[HTML]{000000} 0.05} & \cellcolor[HTML]{FFABA7}{\color[HTML]{000000} 0.09} & \cellcolor[HTML]{FFABA7}{\color[HTML]{000000} 0.27} \\ \cline{2-2} \cline{8-11} 
\multirow{-2}{*}{\cellcolor[HTML]{C0C0C0}Fair-SMOTE} & Race                                                          & \multirow{-2}{*}{\cellcolor[HTML]{FFABA7}{\color[HTML]{000000} 0.71}} & \multirow{-2}{*}{\cellcolor[HTML]{FFFFFF}{\color[HTML]{000000} 0.24}} & \multirow{-2}{*}{\cellcolor[HTML]{FFFFFF}{\color[HTML]{000000} 0.49}} & \multirow{-2}{*}{\cellcolor[HTML]{FFFFFF}{\color[HTML]{000000} 0.75}} & \multirow{-2}{*}{\cellcolor[HTML]{FFABA7}{\color[HTML]{000000} 0.59}} & \cellcolor[HTML]{FFABA7}{\color[HTML]{000000} 0.01} & \cellcolor[HTML]{FFABA7}{\color[HTML]{000000} 0.03} & \cellcolor[HTML]{FFABA7}{\color[HTML]{000000} 0.08} & \cellcolor[HTML]{FFABA7}{\color[HTML]{000000} 0.22} \\ \hline
\end{tabular}
\end{table*}

\begin{RQ}
{\bf RQ2.} Are standard class balancing techniques helpful to reduce bias?
\end{RQ}
For answering RQ2, we chose the most used class balancing technique, which is SMOTE \cite{Chawla_2002}. 
Table \ref{RQ2_3} contains results for three datasets and three different learners (LSR, RF, SVM).  In that table,
for a particular dataset and for a particular performance metric:
\bi
\item
Cells with 
\colorbox{deeppink}{darker backgrounds}  denote treatments that are performing
{\em better} than anything else.
\item
Conversely, cells with a white background denote treatments that are performing {\em worse} than anything else;.
\ei

SMOTE consistently increases bias scores (AOD, EOD, SPD, DI) mean damaging fairness but performs similar/better than ``Default'' in case of performance metrics (as measured by recall, false alarm, precision, accuracy \& F1).
Thus, the answer for RQ2 is ``\textbf{No, standard class balancing techniques are not helpful  since, in their enthusiasm to optimize model performance, they seem to also amplify model bias.}''

\begin{RQ}
{\bf RQ3.} Can Fair-SMOTE reduce bias?
\end{RQ}
Table \ref{RQ2_3} answers this question. Looking at the bias metrics (AOD, EOD, SPD, DI), Fair-SMOTE significantly reduces all the bias scores mean increasing fairness (see the \colorbox{deeppink}{darker colored cells}).

\definecolor{OurPink}{HTML}{FFABA7}

As to the performance measures (recall, false alarm, precision, accuracy,  and F1) it is hard to get a visual summary of the results just by looking at  Table \ref{RQ2_3}. For that purpose,
we turn to rows 1,2,3,4 of  Table \ref{Fair-SMOTE_vs_others}. Based on Scott-Knott tests of \S\ref{subsetion_scott_knott},  these   rows count the number of times Fair-SMOTE wins, losses or ties compared to SMOTE (here, ``tie'' means ``is assigned the same rank by Scott-Knott''). Those counts,  in rows 1,2,3,4  confirm the visual patterns of  Table \ref{RQ2_3}:
\bi
\item 
As to the performance measures (recall, false alarm, precision, accuracy,  and F1), these methods often tie.
\item
But looking at the \colorbox{OurPink}{highlighted} bias metrics results  for AOD, EOD, SPD, \& DI, Fair-SMOTE is clearly performing better than SMOTE.
\ei
Thus, the answer for RQ3 is ``\textbf{Yes, Fair-SMOTE reduces bias significantly and performs much better than SMOTE.}''

\begin{RQ}
{\bf RQ4.} How well does Fair-SMOTE perform compared to the state of the art bias mitigation algorithms?
\end{RQ}
Table \ref{Fair-SMOTE_vs_others} compares Fair-SMOTE against
other tools that try to find and fix bias.
``Default'' shows the off-the-shelf learners;
``Fairway'' is the  Chakraborty et al.~\cite{Chakraborty_2020}
system from FSE'20; and 
OP is the Optimized Pre-processing method from NIPS'17~\cite{NIPS2017_6988}. 
Here, the learner is logistic regression since, for performance measures, LSR has best results in the  Fair-SMOTE results of Table \ref{RQ2_3}.

Rows 5,6,7,8,9,10,11,12 of  Table \ref{Fair-SMOTE_vs_others}
summarize these results.  Measured in terms of bias reduction,
all the methods often tie. But observing the    
\colorbox{OurPink}{highlighted} 
cells in that table, we see that the  Fair-SMOTE  performs much better (in terms of 
recall and F1) than anything else.

Thus, the answer for RQ4 is ``\textbf{Fair-SMOTE performs similar or better than two state of the art bias mitigation algorithms in case of fairness and consistently gives higher recall and F1 score.}'' That means we do not have to compromise performance anymore to achieve fairness. Fair-SMOTE is able to provide both - better fairness and performance. This is the biggest achievement of this work. 
 
\begin{RQ}
{\bf RQ5.} Can Fair-SMOTE reduce bias for more than one protected attribute?
\end{RQ}
In the literature, we are unaware of any work   trying to reduce bias for more than one protected attribute at a time. Typically
what researchers do is 
try to  eliminate bias based on one protected attribute, one  at a time. We experimented on Adult dataset having two protected attributes (sex, race). The idea is to balance the training data with respect to class and two protected attributes. That means we need to find out among the eight ($2^{3}$) possible subgroups which one has the most number of data points. Then we need to make other subgroups of the same size with the one having most number of data points. That is done by generating new data points using Fair-SMOTE. Table \ref{two_protected} shows results that bias is reduced for both the attributes along with higher recall and F1 score. 
Hence, the answer of RQ5 is ``\textbf{Yes, Fair-SMOTE can simultaneously reduce bias for more than one protected attribute.}''

\section{Discussion: Why Fair-SMOTE?}
\label{discussion}
Here we discuss what makes Fair-SMOTE unique and more useful than prior works in the fairness domain. 

\textbf{Combination:} This is only the second SE work (after Fairway \cite{Chakraborty_2020}) in fairness domain where primary focus is bias mitigation. There are a lot of papers in ML domain where various techniques are provided to remove bias from model behavior. Still, when it comes to applicability, we see software practitioners find ML fairness as a complicated topic. Because finding bias, explaining bias, and removing bias have been treated as separate problems and thus created more confusion. That's where this work makes a significant contribution by combining all of them together. We first find data imbalance and improperly labeled data points (by situation testing) and then use oversampling to balance the data and remove improperly labeled data points. As an outcome, we generate fair results.

  
\textbf{Uncompromising:}  Our framework improves fairness scores along with F1 score and recall. It does not damage accuracy and precision much also. That said, unlike much prior work~\cite{NIPS2017_6988,Kamiran2012,10.1007/978-3-642-33486-3_3,zhang2018mitigating,Kamiran:2018:ERO:3165328.3165686,hardt2016equality}, we can do bias mitigation {\em without} compromising predictive performance. We attribute our success in this regard to our sampling policy. None of our mutators damage the associations between attributes. Rather, we just carefully resample the data to avoid certain hard cases (where the training data can only see a few examples of each kind of row).
  
\textbf{Group \& Individual:}  Our data balancing approach takes care of \textit{group fairness} where goal is based on the protected attribute, privileged and unprivileged groups will be treated similarly. Our situation testing method takes care of \textit{individual fairness} where goal is to provide similar outcomes to similar individuals.
    
 \textbf{Generality:}  We entirely focused on data to find and remove bias. There are works where optimization tricks have been used while model training to remove bias \cite{zhang2018mitigating,10.1007/978-3-642-33486-3_3}. These works are model specific and most of the time combine with internal model logic. However, Fair-SMOTE does not require access to inside model logic. Thus it is much more general as it can be used for any kind of model. In this work, we used three simple models and got promising results. In future, we will explore deep learning models.
     
 \textbf{Versatility:}  We used Fair-SMOTE for only classification datasets here. However, the core idea of Fair-SMOTE is keeping equal proportion of all the protected groups in the training data. We believe the same approach can be applied to regression problems. In future we would like to explore that. Besides, the same approach can be applied for image data (face recognition) to train model with equal proportion of white and black faces so that model does not show racial bias. That means Fair-SMOTE can be easily adopted by other domains to solve bias issues caused by data imbalance.
  
\section{Threats to validity}
\label{threats}

\textbf{Sampling Bias} - As per our knowledge, this is the most extensive fairness study using 10 real-world datasets and 3 classification models. Still, conclusions may change a bit if other datasets and models are used.\\
\textbf{Evaluation Bias} - We used the four most popular fairness metrics in this study. Prior works \cite{Chakraborty_2020,hardt2016equality,10.1007/978-3-642-33486-3_3} only used two or three metrics although IBM AIF360 contains more than 50 metrics.  In future, we will explore more evaluation criteria.\\
\textbf{Internal Validity} - Where prior researchers~\cite{zhang2016causal,loftus2018causal} focused on attributes to find causes of bias, we concentrated on data distribution and labels. However, there could be some other reasons. A recent Amazon paper comments on some other reasons such as objective function bias, homogenization bias \cite{Das2020fairML}. We could not validate these biases in our datasets as it was out of scope. In future, we would like to explore those reasons if industry datasets become available. \\
\textbf{External Validity} - This work is based on binary classification and tabular data which are very common in AI software. We are currently working on extending it to regression models. In future work, we would extend this work to other domains such as text mining and image processing.

\section{conclusion}
\label{conclusion}
This paper has tested the Fair-SMOTE tactic for mitigating bias in ML software. Fair-SMOTE assumes  the  root causes of bias are the prior decisions that control
(a)~what data was collected and (b)~the labels assigned to the data.
Fair-SMOTE: 
\begin{quote}
{\em 
(A)~Removes biased labels; and (B)~rebalances internal distributions such that they are equal based on class and sensitive attributes.}
\end{quote}
As seen above, Fair-SMOTE was just as effective at bias mitigation as two other state-of-the-art algorithms~\cite{Chakraborty_2020,NIPS2017_6988} and more effective in terms of achieving higher performance (measured in terms of recall and F1). Also, Fair-SMOTE runs 220\% faster (median value across ten data sets) than Chakraborty et.al~\cite{Chakraborty_2020}.

\noindent
Based on the above, we offer three conclusions:
\be
\item
We can recommend Fair-SMOTE for bias mitigation.
\item
 We can  reject the  pessimism of Berk et al.~\cite{berk2017fairness} who, previously,  had been worried that the cost of fairness was a reduction in learner performance.
\item
More generally, rather than blindly applying some optimization methods it can be better to:
\bi
\item Reflect on the domain; 
\item Use insights from that reflection to guide improvements in that domain. 
\ei
\ee
\section*{Acknowledgements}
The work was partially funded by LAS and NSF grant \#1908762.
\balance

\bibliographystyle{IEEEtran} 

\bibliography{main}

\end{document}